\documentclass{article}

\usepackage{microtype}
\usepackage{graphicx}
\usepackage{subfigure}
\usepackage{booktabs}
\usepackage{array}
\usepackage{amsmath}
\usepackage{pifont}
\usepackage{hyperref}
\usepackage{etoolbox}
\usepackage{units}
\usepackage{multirow}
\usepackage[most]{tcolorbox}
\usepackage{listings}

\usepackage{xcolor} \def\gray#1{\textcolor{gray}{(#1)}}

\newcommand{\parspace}{\vspace*{0.7mm}}

\usepackage[accepted]{icml2021}

\icmltitlerunning{Machine translation decoding beyond beam search}

\begin{document}
\definecolor{dkgreen}{rgb}{0,0.6,0}
\definecolor{gray}{rgb}{0.5,0.5,0.5}
\definecolor{mauve}{rgb}{0.58,0,0.82}

\twocolumn[
\icmltitle{Machine Translation Decoding beyond Beam Search}

\icmlsetsymbol{equal}{*}

\begin{icmlauthorlist}
\icmlauthor{R\'emi Leblond}{dm}
\icmlauthor{Jean-Baptiste Alayrac}{dm}
\icmlauthor{Laurent Sifre}{dm}
\icmlauthor{Miruna Pislar}{dm}
\icmlauthor{Jean-Baptiste Lespiau}{dm}
\icmlauthor{Ioannis Antonoglou}{dm}
\icmlauthor{Karen Simonyan}{dm}
\icmlauthor{Oriol Vinyals}{dm}
\end{icmlauthorlist}

\icmlaffiliation{dm}{DeepMind.}

\icmlcorrespondingauthor{Rémi Leblond}{remileblond@google.com}

\icmlkeywords{Machine Translation, MCTS, Decoding}

\vskip 0.3in
]

\printAffiliationsAndNotice{}

\begin{abstract}
Beam search is the go-to method for decoding auto-regressive machine translation models.
While it yields consistent improvements in terms of BLEU, it is only concerned with finding outputs with high model likelihood, and is thus agnostic to whatever end metric or score practitioners care about.
Our aim is to establish whether beam search can be replaced by a more powerful metric-driven search technique.
To this end, we explore numerous decoding algorithms, including some which rely on a value function parameterised by a neural network, and report results on a variety of metrics.
Notably, we introduce a Monte-Carlo Tree Search (MCTS) based method and showcase its competitiveness.
We provide a blueprint for how to use MCTS fruitfully in language applications, which opens promising future directions.
We find that which algorithm is best heavily depends on the characteristics of the goal metric; we believe that our extensive experiments and analysis will inform further research in this area.
\end{abstract}

\section{Introduction}
Sequence to sequence model decoding remains something of a paradox.
The most widely adopted training method for these models is maximum likelihood estimation (MLE), which aims at maximising the probability of the ground truth outputs provided in the training datasets.
Consequently, decoding from MLE-trained models is done by trying to find the output to which the model assigns maximum likelihood.
Unfortunately, as models usually predict tokens one by one, exact search is not feasible in the general case and practitioners resort to heuristic mechanisms instead.

The most popular of these heuristics is beam search~\citep{reddy}, which maintains several hypotheses in parallel and is guaranteed to find a more likely output than the more basic greedy decoding.
This approach has some obvious flaws: for one, it is completely agnostic to the actual metrics (or scores) practitioners actually want to optimise.

Even more crucially, in most cases beam search fails at the one thing it is supposed to do: finding the optimal output sequence (w.r.t the model), as shown by~\citet{StahlbergByrne2019nmt}.
Also alarming are the findings of~\citet{welleck2020consistency}, proving that traditional search mechanisms can yield infinite-length outputs, to which the model assigns zero probability.
Interestingly, the use of likelihood as a training objective has a spectacular side-effect: it causes trained models to have an inordinate fondness for empty outputs.
By using exact search on the output likelihood in machine translation,~\citet{StahlbergByrne2019nmt} show that in more than half of cases the highest scoring output according to the model is the empty sentence!

All told, we rely on models placing a surprising emphasis on empty outputs, and on a decoding mechanism which usually fails to find optimal outputs; and both ignore the relevant metrics.
One can then justifiably wonder why we observe impressive MT results.
~\citet{StahlbergByrne2019nmt} provide an apparently paradoxical explanation: it is precisely because the decoding mechanisms are imperfect that models produce outputs of high quality.
~\citet{meister-etal-2020-beam} elaborate on this assumption; they show that beam search optimises for a slightly modified likelihood objective, promoting uniform distribution probability inside sentences.

This state of affairs seems highly unsatisfactory.
While a whole body of work has been devoted to alleviating these issues, most approaches have been concerned with training~\citep{bengio2015scheduledsampling,Ranzato2016b,Shen2016mrt,Norouzi2016b,Bahdanau2017actorcritic,edunov2018splosses,leblond2018searnn}, or making the search mechanism differentiable~\cite{collobert2019fdbs}.
These have resulted in performance increase, but they still rely on likelihood as an objective for decoding.
Further,~\citet{choshen2019rlweakness} shows that performance improvements using RL are limited and poorly understood.

In this paper, we focus instead on contrasting the performance of beam search to alternative decoding algorithms aimed at optimising various metrics of interest directly, via a value function (or the metric itself when available).
Notably, we experiment with variants of the powerful Monte Carlo Tree Search (MCTS)~\citep{Coulom2006mcts,Kocsis2006mcts} mechanism, which has a proven track record in other sequential applications~\cite{browne2012mcts,silver2017alphago}.
We investigate whether, by optimising the metric of interest at test time, one can obtain improved performance compared to likelihood-based approaches, and whether performance scales with the amount of computation -- as opposed to that of beam search which has been shown to degrade with large beam sizes~\citep{cohen2019bs}.

We concentrate on machine translation (MT), an emblematic and well-studied sequence to sequence task, which comes with readily available data and well-defined benchmarks.

\parspace
\noindent
\textbf{Contributions.}
(i) We distinguish two different types of metrics: \emph{privileged} scores, which rely on ground truth translations, in contrast to \emph{unprivileged} ones.
We design a new score, Multilingual BERTScore, as an imperfect but illustrative example of the latter.
(ii) We introduce several new decoding algorithms, detailing their implementation and how best to use them for MT.
In particular, we provide a blueprint for how to use MCTS profitably in NLP (as well as pseudocode for a batched Numpy-based~\citep{harris2020numpy} implementation), which opens the door for many exciting applications.
(iii) We run extensive experiments to study the performance of decoding mechanisms for different metrics.
We show that beam search is the best option only for \emph{privileged} metrics. 
For those, value-based alternatives falter as the value problem is too hard  -- since it ultimately relies on reconstructing hidden information.
For unprivileged scores, beam search is outperformed by its competitors, including MCTS.

\parspace
\noindent
\textbf{Outline.}
We go over the related work thoroughly in Section~\ref{relatedwork}.
In Section~\ref{setup}, we contrast several types of metrics, and introduce illustrative examples.
We review beam search and introduce alternative algorithms in Section~\ref{makingitwork}.
We explain how we train the required value function for value-based methods in Section~\ref{value}.
In Section~\ref{results} we introduce necessary architecture adaptations for inference-intensive applications and then go over experimental details and results.
Finally, we discuss our results and their limitations as well as possible next steps in Section~\ref{discussion}.

\section{Related Work}\label{relatedwork}
Incremental models for sequence generation typically output more coherent sequences, as each token prediction takes into account its predecessors~\citep{gu2018nat}.
However, this gain comes at a cost in terms of tractability: finding the sequence with maximum probability according to the model -- $\mathrm{argmax}_{y \in \mathcal{Y}} {\pi(y | x)}$ -- becomes a search problem over the combinatorial space $\mathcal{Y}$.
Given the size of the (token) action space $\mathcal{A}$, exact search appears out of the realm of possibility.
So we have to resort to incremental prediction; but then how do we pick individual tokens, without knowing how these choices will impact the likelihood of the final sequence?
We start by describing the three most widely used methods, which all pick tokens one by one from left to right.

The \textit{sampling} method predicts tokens by directly sampling from the model policy $\pi(y_{t+1} | x, y_1 ... y_t)$, computed via a softmax operator applied to the model logits~\citep{ackley1985sampling} -- possibly after applying a temperature parameter.
The \textit{greedy search} method incrementally picks the tokens with highest probability according to the model.
This inexpensive approach can be seen as a special case of the sampling method, with very low temperature.
Finally, \textit{beam search} maintains a beam of $k$ possible translations, updating them incrementally by ranking their extensions via the model likelihood.
While $k$ times more expensive than the previous approaches, beam search has stood the test of time, resulting in steady performance improvements on MT tasks.

Building on these methods, a number of improvements have been proposed.
\citet{welleck2019outoforder} explore out-of-order decoding, where the model additionally learns the order in which to decode tokens.
This provides benefits in a variety of tasks, but unfortunately not MT.
\citet{Wang2020bslookahead} use look-ahead in the beam search to take into account future likelihood, which yields improvements on low-data tasks, but again does not outperform beam search on MT.
\citet{meister2020bestfirst} speeds up beam search for monotonous scores.

Several works focus on studying the interplay between the incremental models and beam search.
\citet{cohen2019bs} shows that performance is not monotonically increasing with beam size, but degrades after a fairly small value of $k$.
\citet{StahlbergByrne2019nmt} devise a clever exact search mechanism, relying on the fact that likelihoods are monotonically decreasing with size.
While still prohibitively expensive, this approach underlines several key facts.
First, beam search does not recover $\mathrm{argmax}_{y \in \mathcal{Y}} {\pi(y | x)}$ in a most cases, even with increased computational budget.
Second, $\mathrm{argmax}_{y \in \mathcal{Y}} {\pi(y | x)}$ is the empty sequence more than half the time in MT.
\citet{Eikema2020MBR} propose an interesting explanation for this observation: while models are good at spreading probability mass over a large quantity of acceptable outputs, they are unable to effectively pick the best one.
Indeed, the mode of the distribution might even be disjoint from the area where the models assigns the majority of probability mass.
They propose \textit{minimum Bayes risk} decoding, which leverages the whole distribution rather than only its mode, and can outperform vanilla beam search in low-resource scenarios.

A large body of work has been dedicated to improving sampling diversity, which plays a key role in many NLP applications -- though not usually in machine translation.
\citet{fan2018topksampling} propose only sampling from the top $k$ tokens according to the policy to avoid sampling from the tail of the distribution.
\citet{holtzman2019nucleus} adopt a similar approach, but instead of fixing $k$ they fix $p$, the size of the `nucleus' of the distribution from which sampling is allowed to select tokens.
This performs better on open-ended tasks.
\citet{kool2019stochasticbeams} propose a search mechanism in-between sampling and beam search, which produces provably unique samples by leveraging the Gumbel-Max trick~\citep{gumbel1954max}.
\citet{yu2020mctswmt} use a different, much more expensive flavor of MCTS to add diverse samples to a larger NMT system: instead of relying on direct value estimation, they rely on (expensive) rollouts to estimate node values.

Finally, the most pertinent approach to optimise various metrics -- and most closely related to our proposed MCTS decoding -- is value-guided beam search, as developed by~\citet{he2017decodingvalue,ren2017captioning} for MT and image captioning.
Contrary to all other methods presented in this section, this approach does not solely rely on model likelihood.
In both papers a value network -- estimating the eventual score from an unfinished sample -- is trained in addition to the policy network.
Then instead of following the likelihood to select the hypotheses on the beam, one uses a linear combination of the policy logits and the value.
This approach has shown improved performance compared to vanilla beam search; notably, it is less sensitive to the chosen beam size.
While this method uses the value exclusively for one-step look-aheads, MCTS can be leveraged to explore further in the future.
Additionally, it requires evaluating the value score of all tokens at each step, which can be prohibitively expensive if the action space is big (in MT, one routinely uses vocabularies of size larger than 30000).

\section{Machine Translation metrics}\label{setup}
There are two main evaluation strategies for MT outputs.
The first one crucially relies on having access to a held-out test set of high quality (input, output) sentence pairs $(x, y_x)_{x \in \mathcal{X}}$.
One can then compute a monolingual similarity score between the system's outputs $(\hat{y}_x)_{x \in \mathcal{X}}$ and the ground truth outputs $(y_x)_{x \in \mathcal{X}}$.
Common metrics include BLEU~\citep{papineni2002bleu}, METEOR~\citep{denkowski2014meteor} which takes into account synonyms; or BERTScore~\citep{zhang2019bertscore}.
We refer to this type of metrics as \emph{privileged}, as they require access to ground truth translations.

The second is concerned with assessing translation quality for source sentences for which one does not have reference translations.
To determine whether machine-generated outputs are accurate enough or require human modification, one relies on multilingual quality estimation metrics~\citep{specia2018qe}.
These do not rely on ground truth sequences; instead comparing produced samples to sources sentences directly.
Expert human evaluation is perhaps the most relevant such score, but many automated alternatives exist~\citep{martins2017qe}; see e.g.~\citet{bhattacharyya2021energybased}.
We refer to these metrics as \emph{unprivileged}.

Privileged metrics provide high quality evaluation signal, and are well-suited to comparing average model performance (trusting that results on the unseen test set generalise to other domains of interest).
However, they rely on the quality of the test set translations (which are usually unique, hence somewhat arbitrary), and cannot be used to evaluate the quality of models' prediction for specific unseen inputs.
In contrast, unprivileged metrics are harder to access or approximate but can be used without ground truth translations.

We use two privileged metrics in our experiments: BLEU and BERTScore.
We introduce another, unprivileged metric: Multilingual BERTScore.
Note that while a translation model likelihood can be considered an unprivileged metric, it comes with the unusual property that it is decomposable.
We thus treat it as a special case.

\noindent
\textbf{BLEU~\citep{papineni2002bleu}.}
The BLEU score computes modified precisions for n-grams (typically with n ranging from 1 to 4) between a corpus of candidate sentences and a reference corpus.
These precisions are then averaged geometrically, and multiplied by a brevity penalty.
This metric is meant to be used at the corpus level; it is unstable at the sentence level.
It is the de facto golden standard for comparing MT algorithms, though as it crucially relies on access to a dataset of reference translations, it is not available to assess translation quality at decoding time.

\parspace
\noindent
\textbf{BERTScore~\citep{zhang2019bertscore}.}
By contrast, BERTScore is a sentence-level metric to compare a candidate sentence to a reference translation.
It relies on several consecutive steps: first, computing contextual embeddings for each token in both sentences with a shared BERT~\citep{devlin2019bert} model; second, computing all pairwise cosine similarities between embeddings of the two sentences; third, greedily aligning tokens based on these similarities; finally, averaging the similarities of the aligned tokens.
Compared to BLEU, BERTScore is found to correlate slightly better with human judgement.
Importantly for decoding purposes, it is a sentence-level metric (which is averaged to produce a corpus-level statistic).

\parspace
\noindent
\textbf{Multilingual BERTScore.}
While BERTScore is designed as a monolingual metric, we repurpose it as a multilingual one by using it to compare a candidate to its \emph{source sentence} (instead of reference translation).
Both sentences are in different languages, but this is fine as long as the underlying BERT model is itself multilingual.\footnote{We use the multilingual, 12-layer, 768 hidden dimensionality BERT model available at \url{https://github.com/google-research/bert} (23/11/2018 entry).}
We call this new metric Multilingual BERTScore.
Its performance relies on the underlying BERT model's ability to map related tokens in different languages to similar embeddings.
Because of the one-to-one nature of the alignment phase, we expect it to score more highly translation pairs that have a one-to-one token correspondence, rather than syntactically different pairs.
We thus expect it to make sense for pairs of syntactically similar languages.
We stress that we do not advocate widespread adoption of this imperfect score for MT; we consider it however a convenient illustrative example of an unprivileged metric.
In practice, we observe that it behaves reasonably for our two evaluation language pairs (WMT '14 English/German and English/French) as shown in Table~\ref{table:MLBERTScore}.
Interestingly, it scores trained model outputs higher than ground truth outputs.
We hypothesise that the former follow the source sentence more closely than the latter.
\begin{table}[]
\footnotesize
\centering
\begin{tabular}{lcccc}
\toprule
\textbf{Dataset} & \textbf{Random} & \textbf{GT} & \textbf{Greedy} & \textbf{Beam search} \\ \midrule
ENDE             & 68.13           & 81.36       & 83.02           & 83.40                 \\
ENFR             & 68.00              & 82.83       & 83.85           & 84.00                \\  \bottomrule
\end{tabular}
\caption{Multilingual BERTScore for random text, ground truth sentences (GT), greedy and beam search decoding from a supervised model (guided by likelihood). We see that (i) random translation scores are lower than those of reference translations or that of supervised policy samples (though it is not 0, as for standard BERTScore, as both rely on continuous embeddings similarities); (ii) the scores are smaller than the corresponding monolingual BERTScore (87.88 and 90.55 for greedy decoding from a supervised model, respectively) and (iii) beam search outputs outperform greedy outputs consistently (as is the case for BLEU).}
\label{table:MLBERTScore}
\end{table}

\section{Decoding algorithms}\label{makingitwork}
In this section, we go over the details of each algorithm, and its adaptations to better suit privileged or unprivileged metrics.
We separate them in three categories: (i) algorithms based on likelihood maximisation, (ii) value-based mechanisms which rely on approximating metrics via a value function, and (iii) ranking-based methods which access the metrics directly and pick the highest-scoring example out of a pool of finished candidates.
Of course, ranking-based methods are only usable for unprivileged metrics, as privileged metrics are not computable at test time.
We provide a high-level comparison of all algorithms in Table~\ref{table:algo}.

\subsection{Likelihood-based decoding}

\noindent
\textbf{Greedy decoding (GD)} is our first baseline; it consists in picking the token with maximum likelihood at each step.

\noindent
\textbf{Beam search (BS)} maintains a beam of $k$ possible translations prefixes at each time step $t$, $(p^i_{t})_{i=1}^k$.
Prefixes are updated incrementally as follows: for each prefix $p^i_{t}$ one adds each of the corresponding $k$ most probable tokens (given $p^i_{t}$), resulting in at most $k \times k$ new prefixes of size increased by 1.
Then among these the $k$ prefixes with the highest likelihood are selected, thus obtaining $(p^i_{t + 1})_{i=1}^k$.
This method aims at optimising likelihood, and is agnostic to any metric of interest.
It is therefore at a disadvantage if we change the objective of the search.
Consequently, we also study the performance of value- or score-based variants.

\subsection{Value-based decoding}
To motivate the introduction of value functions in our decoding mechanisms, it is helpful to understand how machine translation can be construed of as a Reinforcement Learning task, with an underlying Markov Decision Process (MDP).
In MT, we work with a vocabulary $\mathcal{V}$ of tokens, and a dataset contains pairs of sentences $(x, y_x)$ where $x, y_x \in \mathcal{V}^{+}$.
We can define a somewhat trivial MDP, where:
\begin{itemize}
    \item the states consist in a pair containing a source sentence $x \in \mathcal{V}^{+}$ and a sample in construction $\hat y_1... \hat y_t \in \mathcal{V}^{+}$,
    \item the action space $\mathcal{A}$ is the output vocabulary $\mathcal{V}$ (taking an action means adding a specific token to the sample),
    \item the transitions are deterministic: picking token $\hat y_{t+1} \in \mathcal{A}$ in state $s_t = (x, \hat y_1...\hat y_t)$ leads to the unique possible successor state $s_{t+1} = (x, \hat y_1...\hat y_t \hat y_{t+1})$,
    \item the reward is $0$ for any non-terminal state; for terminal states, it is $m(y_x, \hat y)$ for privileged metrics and $m(x, \hat y)$ for unprivileged metrics of interest. 
    Entering a terminal state is done by picking a special \texttt{<EOS>} token.
\end{itemize}
A \emph{value function} $v$ for a policy $\pi$ approximates the final score one might expect to obtain, starting from a non-terminal state $s$, and following $\pi$ thereafter.
It thus provides \emph{forward-looking} guidance \emph{during} decoding, as opposed to likelihood (accessible during decoding but myopic) or a score (only computable on finished sentences).

\parspace
\noindent
\textbf{Value-guided beam search (VGBS)}, as developed by~\citet{he2017decodingvalue,ren2017captioning}, augments the decision mechanism in beam search (when picking the top $k$ prefixes amongst $k \times k$ candidates) with a value network $v$.
The internal score is a linear combination between the (length-normalised) log-likelihood of a prefix and its value approximated by the value network, with a contribution factor $\alpha$: $bs(s_t, a_t) = \frac{\alpha}{t} \log\big(\pi(s_t, a_t)\big) + (1 - \alpha) v(s_t, a_t)$.\footnote{Using the logarithm of the value instead, as in~\citet{he2017decodingvalue}, yields no practical gains. So we opt for the simpler formulation.}
Note that this method does not use the score; thus it is applicable to both unprivileged and privileged metrics.

\parspace
\noindent
\textbf{Value-guided MCTS (V-MCTS)} indicates the version of Monte Carlo Tree Search as used by~\citet{silver2017alphago}.
This search method combines both a policy $\pi$ and a value network $v$. 
For every decoding step, a fixed budget of simulations is allocated to build a tree of possible future trajectories.
Each simulation consists in 3 steps:
\begin{itemize}
    \item selection: recursively picking children nodes according to the pUCT formula, starting at the root and until reaching an unopened (i.e. not expanded yet) node $s_o$:
    \begin{equation*}
    \begin{aligned}
        a &= \underset{{a \in \mathcal{A}}}{\mathrm{argmax}}\Bigg(Q(s, a) + c_{\text{puct}} \pi_{\tau}(a | s) \frac{\sqrt{\sum_b N(s, b)}}{1 + N(s, a)}\Bigg)
    \end{aligned}
    \end{equation*}
    where $Q(s, a)$ is a statistic representing the value of taking action $a$ in state $s$, updated online during the search, $c_{\text{puct}}$ is a tunable constant, $\tau$ is a temperature parameter applied to the policy $\pi_\tau (a | s) = \nicefrac{\pi(a | s)^{1/\tau}}{\sum_b \pi(b | s)^{1/\tau}}$ and $N(s, a)$ is the number of times action $a$ has been chosen from state $s$ while building the tree (also called visit count);
    \item expansion: opening the selected node $s_o$ by computing the policy $\pi(a | s_o)_{a \in \mathcal{A}}$ at the associated new state, as well as the value $v(s_o)$;
    \item backup: updating the $Q$ statistics encountered during the tree traversal leading to $s_o$ via an aggregation mechanism (such as \emph{averaging} the previous statistic with $v(s_o)$, or taking their \emph{maximum}: ${Q \leftarrow \mathrm{max}(Q, v(s_o))}$).
\end{itemize}
Once the tree is finished, the decision for the current decoding step is made according to the statistics of the root's children nodes.
A popular option consists in picking the root child with the \emph{most visit counts}, but one may also select the one with \emph{maximum aggregated value} instead.

While it is customary to allow MCTS to use the score directly when encountering a terminal state, we opt for a pure value implementation instead (i.e. using the value instead of the score on terminal nodes).
This makes V-MCTS applicable to privileged metrics, which it wouldn't be otherwise.

One of the keys to successful MCTS performance is properly balancing the breadth and depth of the exploratory trees.
We found two adaptations to be helpful. 
First, we used an adaptive value scale as described by~\citet[Appendix B]{schrittwieser2020muzero}:
in the selection phase, we rescale $Q(s, a)$ in the $[0, 1]$ interval by replacing it with $\frac{Q(s, a) - \min Q}{\max Q - \min Q}$, where $\min Q$ and $\max Q$ correspond to the minimum and maximum value observed in the tree, updated online.
Second, we tune the logits temperature $\tau$ jointly with the $c_{\text{puct}}$ hyper-parameter.

\subsection{Reranking-based decoding}
Value-driven decoding methods are well-suited to optimise metrics which we cannot evaluate at test time, such as privileged metrics.
One might also prefer them for especially expensive unprivileged metrics, e.g. expert human evaluation.
For tractable unprivileged metrics though, we can directly compute the scores of finished candidate sentences, without having to resort to approximation.
We study two specific decoding mechanisms that take advantage of this option.

\parspace
\noindent
\textbf{Sampling and reranking (S+R, S+RV)} simply consists in sampling a fixed number of finished candidate sentences $\hat y^1, ..., \hat y^n$ from the policy (with a carefully tuned temperature applied to its logits), scoring all of them and picking the highest-performing one: $\mathrm{argmax}_{i=1}^n m(x, \hat y^i)$.
To measure the loss of performance associated with using a value, we also introduce a variant, S+RV that ranks candidates according to the value (rather than the score).

\parspace
\noindent
\textbf{MCTS with rollouts (MCTS+Roll)} is a variant of V-MCTS where we replace the value approximation for a given node $s$ by a more expensive one based off the actual score.
From $s$, we perform a greedy rollout (w.r.t. the policy $\pi$) until we arrive at a terminal node $s_T$.
We then compute the score with the finished sample and the source as inputs, use this scalar as the value of node $s$, and continue as in V-MCTS.
Of course greedy rollouts are expensive in MT, so this method is not directly comparable to V-MCTS.
It is however useful as a proof of concept which enables us to measure how much performance we lose by relying on a value function rather than directly on the score.

\begin{table}[]
\centering
\footnotesize
\begin{tabular}{lcc}
\toprule
\multicolumn{1}{l}{} & \multicolumn{1}{l}{Uses a value} & \multicolumn{1}{l}{Uses the score directly} \\ \midrule
Greedy               &              \ding{55}                    &               \ding{55}                     \\
Beam Search          &              \ding{55}                    &               \ding{55}                     \\
VGBS                 &              \ding{51}                    &                 \ding{55}                   \\
V-MCTS               &              \ding{51}                    &                 \ding{55}                   \\
S + RV               &              \ding{51}                    &                 \ding{55}                   \\
S + R                &              \ding{55}                    &                  \ding{51}                  \\
MCTS + Roll          &              \ding{55}                    &                  \ding{51}  \\  
\bottomrule
\end{tabular}
\vspace*{-2mm}
\caption{Decoding algorithms characteristics.}
\label{table:algo}
\vspace*{-3mm}
\end{table}

\section{Training a value network}\label{value}
Several of the algorithms we detail in Section~\ref{makingitwork} make use of a value network.
We train such models in several steps.
\begin{itemize}
    \item First, we train a plain supervised policy model $\pi_{\text{sup}}$ on our bilingual datasets.
    \item Second, we update each data item $(x, y_x)$, which contains a source $x$ and a reference sentence $y_x$, by replacing $y_x$ with a sample $\hat y_x$ obtained via greedy decoding\footnote{We also tried sampling one or more sentences instead, but did not detect any improvements from doing so.} from our trained policy $\pi_{\text{sup}}$, and adding a score $m$ comparing either $\hat y_x$ to $y_x$ (for privileged metrics) or $\hat y_x$ to $x$ (for unprivileged metrics).
    \item Finally, we train a dual-headed network on the augmented dataset, with a shared transformer encoder-decoder torso~\citep{vaswani2017transformer} taking source $x$ and sample $\hat y_x$ as inputs, and two heads, one predicting the policy $\pi_d$ and the other the value $v$. This approach provides a powerful regulariser for the value, greatly reducing its tendency to overfitting~\citep{silver2017alphago}. 
\end{itemize}
The second step is mandatory to obtain a score distribution to train the value model on, in the case of privileged metrics.
Indeed, the scores of the optimal supervised policy are all perfect (comparing $y_x$ to $y_x$), thus uninformative, making it impossible to train a value network on.
Relying on a sample rather than on the ground truth sentence to compute the score has another advantage: the samples follow the policy $\pi_{\text{sup}}$ so the value will be the one associated with a trained policy, as the one we use during decoding, rather than with the optimal supervised policy.

\parspace
\noindent
\textbf{Losses.} We train the policy by minimising its Kullback-Leibler divergence with the initial supervised policy $\pi_{\text{sup}}$: $\mathcal{L}_{\pi} = D_{\text{KL}}(\pi || \pi_{\text{sup}})$.
We reframe the value regression problem as classification by discretising the score interval into buckets.
We emulate training our value function on unfinished samples by adding a value loss term at every step, and reusing the transformer decoder causality mask.

\parspace
\noindent
\textbf{The trouble with privileged metrics.}
In practice, we find that learning a value function for privileged metrics (such as BLEU or BERTScore) is difficult.
To understand why, we run an ablation to distinguish between the three subtasks a value function must perform: (i) approximate the score, (ii) predict the end of a trajectory from an unfinished prefix, and (iii) assess the translation quality of a pair of finished sentences in different languages.
To separate concerns, we run the following experiments: for (i), we train our network to predict BLEU given ground truth targets (rather than source sentences) and finished samples (instead of prefixes).
For (ii), we give the network ground truth targets and \emph{unfinished samples}. 
Finally, for (iii), we give the network source sentences and finished samples (thus removing the need to predict the future of trajectories).
We observe that: the error is very low for (i); higher, but significantly improved over the full setup for (ii); and surprisingly, roughly identical to the full setup for (iii).
Thus the real difficulty lies in (iii).
\begin{table}[]
\footnotesize
\centering
\begin{tabular}{lll}
\toprule
\textbf{}                                   & \multicolumn{2}{c}{\textbf{Decoder inputs}}                             \\ \cmidrule{2-3} 
\multicolumn{1}{l}{\textbf{Encoder inputs}} & \multicolumn{1}{l}{Unfinished sample} & \multicolumn{1}{l}{Finished sample} \\ \midrule
Source                              & 0.112 (full setup)                                  & 0.111 (iii)                                    \\
Ground Truth Target                              & 0.065 (ii)                                  &  0.002 (i)                                   \\ \bottomrule
\end{tabular}
\vspace*{-2mm}
\caption{$\ell_1$ error of BLEU value networks trained on different encoder inputs and outputs on our sample dataset based on WMT14 ENDE. Scores (and hence $\ell_1$ error) are between 0 and 1.}
\vspace*{-3mm}
\end{table}

One possible explanation for this result is that the value network is missing a key input.
Indeed, in the case of privileged metrics, the score is computed between a sample $\hat y_x$ and a ground truth reference $y_x$; but the value network only has access to the source sentence $x$ and a prefix of $\hat y_x$.
Thus before it can compute a precise score approximation, it first has to infer $y_x$ from $x$.
But of course, inferring $y_x$ from $x$ is exactly the original machine translation problem, which makes the value problem empirically harder than its policy counterpart on our dataset.

\parspace
\noindent
\textbf{Hybrid architecture for privileged metrics.}
We see that a ``cheating'' value network (as in (ii)) performs strongly.
Unfortunately, we cannot allow our model to cheat at test time.
However, we can still leverage privileged information at training time through representation shaping, by distilling a cheating value network into a non-cheating one.
We propose a new training paradigm, where we call the transformer model twice per step.
The first call is the regular pathway, with source sentence fed to the encoder and sample to the decoder.
The second call is the ``cheating'' pathway, with ground truth reference fed to the encoder and sample to the decoder.
We apply the policy and value losses described earlier in this section to the regular pathway outputs.
To the cheating pathway outputs, we apply only a value loss.
Finally, we add a simple $\ell_2$ loss between the final layers of both pathways.
The idea is to use the powerful cheating representation to help the weaker regular representation.
An illustration can be found in Appendix~\ref{apx:hybrid}.

At inference time, we only compute the regular pathway, which does not cheat.
In practice, with careful tuning of the loss hyper-parameters we are able to significantly reduce the gap in performance between this new hybrid model and the cheating one (as in (ii)), so we use this training regimen for our experiments on privileged metrics.

\section{Experiments}\label{results}
We start by detailing our general setup, then report results for all 3 metrics we consider, move on to describe how we tuned all algorithms for best performance, and finally study how they scale with increasing search budget.

\begin{table*}[ht]
\centering
	\resizebox{0.9\linewidth}{!}{
		\renewcommand{\arraystretch}{0.5}
		\footnotesize
		\begin{tabular}{@{}rccccccc@{}}
			\toprule
			& \multicolumn{3}{c}{ENDE}                                                                   & \multicolumn{3}{c}{ENFR}                                                                   \\ \cmidrule(lr){2-4} \cmidrule(l){5-7} 
			Target score    & \multicolumn{1}{c}{BLEU} & \multicolumn{1}{c}{BERTScore} & \multicolumn{1}{c}{MLBERTScore} & \multicolumn{1}{c}{BLEU} & \multicolumn{1}{c}{BERTScore} & \multicolumn{1}{c}{MLBERTscore} \\ \midrule
			Random          &           0              &               69.59                 &                 68.13                &            0              &             70.55                 &             68.00                 \\
			Ground Truth              &          100                &             100.0                  &              81.36                   &           100               &             100.0                  &                82.83                 \\
			~\citet{vaswani2017transformer}, beam search              &          27.30                &             --                  &              --                   &           38.10               &             --                  &                --                 \\ \midrule
			Greedy      &          25.99                &           87.88                    &              83.02                &            38.70              &           90.55                 &              84.22                   \\
			Beam search &            \textbf{27.75}              &             \textbf{88.48}               &               83.40                 &            39.24              &            90.76                   &           84.48                      \\ \midrule
			VGBS            &           27.17               &               88.47               &              \textcolor{blue}{\textbf{85.10}}           &           \textbf{39.33}               &           90.87                  &           85.80                    \\
			S+R (value)     &           26.03               &               88.39                &          84.49                   &            38.67              &            \textbf{90.93}              &        85.68                      \\
			V-MCTS          &           27.47               &               88.45             &      84.97                    &            39.12              &        90.80                    &     \textcolor{blue}{\textbf{86.31}}                         \\ \midrule
			S+R (score)     &           --               &               --                &         85.11                   &            --              &        --                   &    86.12                         \\
			MCTS + rollouts &           --               &               --                &          \textbf{85.76}                   &            --              &        --                      &        \textbf{86.87}                     \\ \bottomrule
		\end{tabular}}
\caption{Comparison of decoding mechanisms on ENDE and ENFR. Top row contains general metric statistics and the original transformer baseline; the second row performance of supervised models with likelihood-based decodings; the third results for value-based algorithms with joint policy/value models (a specific one for each metric); the last one numbers for score-based methods. Best overall performance is in bold; best value-based performance in blue. Beam search performs strongly for privileged metrics, while value-based methods prevail for unprivileged scores. Score-based methods outperform their value-based counterparts, but V-MCTS remains competitive.}
\label{table:privileged}
\vspace*{-3mm}
\end{table*}

\subsection{Experimental setup}
We consider two established machine translation datasets: WMT'14 English to German (ENDE) and WMT'14 English to French (ENFR).
The first dataset contains roughly 4.5 million training sentence pairs, while the second is much bigger with just under 41 million training sentence pairs, which enables us to account for scale in our experiments.
All dev and test sets contains approximately 3000 sentences.

Our joint policy/value model is based on the Transformer encoder-decoder architecture~\citep{vaswani2017transformer}, which is typically used in machine translation studies.
Encoder and decoder have 6 attention blocks, hidden dimensionality 512, 16 heads and our dictionary size is about 32k.
As we test inference-intensive methods, we use a few adaptations (while still matching or outperforming the original transformer models, as shown in Table~\ref{table:privileged}).

First, we use multi-query attention, as defined by~\citet{shazeer2019singlekv}.
Counter-intuitively, the performance bottleneck for small transformer architectures on our hardware of choice, TPUv3, is memory access by a very large margin.
This is driven by the need to store and read keys and values from memory to enable faster, incremental inference.
We reduce this memory footprint by only computing a single set of keys and values per attention block, that we share across all attention heads.
This simple change yields an impressive, almost linear speedup with respect to the number of attention heads. 
While it comes at a small cost in terms of accuracy, this can easily be offset by reallocating the attention weights we removed to the feed-forward layer of the attention blocks.

Second, we use key and value dimensionality 128, rather than the more customary 64, which enables faster inference by removing an expensive padding operation on TPUv3.

Finally, we allow a budget of 50 inferences per token in the sampled solutions for all methods; compared to 1 for greedy decoding, and 4 for beam search.
We use incremental sampling for speed. 
Both the running time and the memory footprint are directly proportional to the amount of inferences for all methods.

\subsection{Main results analysis}

\parspace
\noindent
\textbf{Privileged metrics: BLEU and BERTScore.}
We report our results on privileged metrics in Table~\ref{table:privileged}.
Plain beam search is a strong contender in this setup, often matching or outperforming other methods, while using a fraction of the inference budget (unfortunately performance degrades rapidly with larger beam sizes so we cannot leverage more computational budget).
In this setup, value-based methods struggle to justify their higher complexity and cost.

\parspace
\noindent
\textbf{Value-based algorithms for unprivileged metrics.}
The results for this alternative use case, also presented in Table~\ref{table:privileged}, paint a completely different picture.
We see that while regular beam size obtains a small but consistent improvement, value-guided methods perform significantly better.
Between the latter, MCTS is particularly promising, as its performance scales nicely with the size of the dataset.

We observe that the policies of our joint policy/value models perform slightly worse than their supervised counterparts (see Table~\ref{table:supdis} in Appendix).
If we use the initial supervised model policy in conjunction with the multilingual value (see Table~\ref{table:supvalue} in Appendix), we obtain promising results: notably 40.31 BLEU when optimising MLBERTScore with MCTS on the ENFR dataset -- more than a full BLEU point above the performance of beam search.
From a qualitative point of view, we see a confirmation of our conjecture: multilingual BERTScore is not perfectly aligned with BLEU.
It seems to encourage word-for-word translations, which has a positive effect initially (more consistency between the source and the sample sentences), but ultimately leads to less natural translations if used with enough budget.

\parspace
\noindent
\textbf{Score-based approaches for unprivileged metrics.}
The bottom of Table~\ref{table:privileged} gives results when we allow direct access to our two unprivileged metrics, without having to go through a value approximation.
They reinforce our finding that the choice of algorithm heavily depends on the use case.

Two additional properties stand out.
First, all the methods that access the score directly perform significantly better than their value-guided counterparts.

Second, the purely value-based V-MCTS is competitive with and can even outperform the score-based approach, S+R.
This is promising, as MCTS is more widely applicable (as some scores are expensive to get).
However S+R performs surprisingly well, which may warrant more explorations of sampling methods optimising for diversity (e.g.~\citet{fan2018topksampling,holtzman2019nucleus,kool2019stochasticbeams}).

\subsection{Algorithms tuning and ablations}
We detail in this section which hyper-parameters we tuned (and how) to generate the results reported in Table~\ref{table:privileged}.
Precise ranges and ablations are provided in Appendix~\ref{apx:expdet}.

\parspace
\noindent
\textbf{Beam search.}
We experiment with the beam size, the normalisation constant, and try to apply temperature to the logits.
We found that the best performance was achieved with the default hyper-parameters for most metrics.

\parspace
\noindent
\textbf{Value-guided beam search.}
In this variant, the ranking rule is a linear combination of the log-likelihoods of tokens and their values provided by a neural network.
We tested various value scaling schemes (e.g. taking the logarithm of the value as detailed by~\citet{he2017decodingvalue}), as well as log-likelihood normalisations; we found that using the plain value, with a length normalisation for the log-likelihoods worked best.
We tuned the weight $\alpha$ of the value in the score.
For privileged metrics, $\alpha=0.5$ performed best.
For unprivileged metrics larger weights were preferable ($\alpha=0.9$ for ENDE and $\alpha = 0.95$ for ENFR), which confirms our observation that value functions are of higher quality in this setup.

\parspace
\noindent
\textbf{Value-guided MCTS.}
We find that selecting the best pair of logits temperature and scaling factor $c_{\text{puct}}$ is key for optimising performance.

Specifically for unprivileged metrics, we made two other key decisions.
First, instead of using the weighted average rule in the backup phase (where the value of the newly opened node is used to update the value of all its predecessors), we found that we could use instead the max operator to better effect.
Second, we observed better results when selecting as action (once the full tree is completed) the root child with maximum \emph{aggregate value}, instead of the more customary \emph{visit count}.
Both the generic options are meant to make the search more robust to value outliers, as they average statistics.
We essentially found that for unprivileged metrics, our value approximation was good enough that following the value more aggressively led to improved performance.
The reverse was true on privileged metrics, where our value networks are lower-quality.

\parspace
\noindent
\textbf{Sampling and reranking.}
Given the simplicity of this approach, there really is a single hyper-parameter to tune: the temperature applied to the model likelihoods before sampling.
We do find however, that properly picking this temperature is critical, as the method is sensitive to it: high temperatures ($\tau \geq 1$) leads to nonsensical outputs, while conversely low temperatures ($\tau \leq 0.5)$ lead to almost no diversity in outputs.
On the dev set, our best performing runs use $\tau = 0.75$ for MLBERTScore, but $\tau = 0.25$ for privileged metrics (using S+RV which relies on value approximation).

\parspace
\noindent
\textbf{MCTS + rollouts.}
We find that the adaptations we made for Value-guided MCTS also perform best for this proof-of-concept algorithm.
As the quality of the value estimation is good (though expensive), following the value aggressively also yields the best results.

\subsection{Scaling search with computational budget}
\begin{table}[]
\small
\begin{tabular*}{\textwidth / 2 - \tabcolsep}{lccc@{\extracolsep{5pt}}ccc@{}}
\toprule
Data       & \multicolumn{3}{c}{ENDE} &  \multicolumn{3}{c}{ENFR} \\ \cmidrule{2-4} \cmidrule{5-7} 
Alg.       & VGBS   & S+R   & MCTS  &  VGBS   & S+R   & MCTS  \\ \midrule
1      &  82.82      &   81.77    &    82.82     &   84.22     &   82.94    &    84.22     \\
10     & 84.48       &   84.25    &   84.47      &   85.38      &   85.35    & 85.76     \\
25     & 84.87        &   84.77    &   84.81      &   85.69      &   85.82    &    86.10     \\
50     &   85.10     &   85.10    &   84.97      &   85.80      &   86.12    &    86.31     \\
75     &   85.34     &   85.31    &   85.07      &   85.75      &   86.33    &    86.44     \\
100    &    85.39    &   85.40    &   85.14      &  85.78       &   86.44    &    86.51     \\
200    &    85.64    &   85.69    &   85.27      &    85.77    &  86.76     &    86.62     \\
300    &    85.79    &   85.89    &   85.27      &    85.79    &  86.90     &      86.66  \\ 
\bottomrule
\end{tabular*}
\vspace*{-2mm}
\caption{Comparison of how our methods scale with search budget (from 1 to 100 inferences per token) on MLBERTScore.}
\label{table:scaling}
\vspace*{-4mm}
\end{table}

We study how our decoding algorithms scale with their search computational budget.
We report results on MLBERTScore in Table~\ref{table:scaling} and more detailed numbers in Appendix~\ref{apx:scaling}.
Our findings are fairly unsurprising: the more quality score data algorithms can leverage, the better they scale.
We see that on privileged metrics -- where value networks are hard to train and thus quite imperfect -- performance quickly stops increasing with more computation and start degrading instead.
When using higher quality value networks (in the unprivileged metrics setup), performance increases more steadily with computation (almost everywhere), plateauing rather than degrading.
Finally, when accessing the score directly (for the ranking approaches), the more computation, the better and performance keeps increasing with more inferences.

\section{Discussion}\label{discussion}
We now summarise and contextualise our findings, and discuss potential next steps for optimising relevant MT metrics.

The main takeaway from our experiments is that which algorithm is best depends heavily on the metric to optimise.
This reinforces the notion that one should carefully consider when picking a decoding mechanism for a machine translation pipeline, rather than default to beam search.

Second, we find that optimising privileged metrics (e.g. BLEU) via a value function is surprisingly hard.
While distinguishing large gaps in quality is easier than modelling the policy, discriminating between good candidates is in practice as hard as the policy problem, since a first step required step is to estimating the ground truth sentence.
Indeed, in our experiments we observe relatively low quality value networks, and comparatively little improvements with value-based decoding methods (especially on the small ENDE dataset).
Using a value function to optimise unprivileged metrics is more promising.

Third, we show that MCTS is not only a valid way of decoding for machine translation tasks, but also the best option in some use cases (given some necessary adjustments).
We study its strengths and weaknesses, and demonstrate that its performance is crucially linked to the ease of learning a good value function.
We include pseudo-code for an easily reproducible Numpy implementation in Appendix~\ref{apx:code}. 
All told, we provide a blueprint for how to use MCTS efficiently in NLP with state-of-the-art transformer models.

Finally and somewhat surprisingly, we find that whenever access to the score is possible, the deceivingly simple S+R method performs well.
More experimentation is required to understand why; but at any rate, it should be a strong contender in this specific setup.

\parspace
\noindent
\textbf{Future directions.}
We have shown that optimising for unprivileged metrics is easier than for privileged ones.
The ultimate unprivileged metric for machine translation is human translation assessment.
Thus it seems natural to consider training a score directly from human evaluation of translation pairs, and to later focus on optimising it via MCTS.

Another natural extension is a full-blown RL algorithm; iteratively improving policies via value-guided search and training value functions on search-improved policies, getting closer to the optimal policy and value at each step. 

\section*{Acknowledgements}
We want to thank our colleagues David Silver, Chris Dyer, Wang Ling, Julian Schrittwieser and Thomas Hubert for fruitful conversations, guidance, contributions to the paper and help with technical support.

\bibliography{mcts_decoding}

\begin{thebibliography}{44}
\providecommand{\natexlab}[1]{#1}
\providecommand{\url}[1]{\texttt{#1}}
\expandafter\ifx\csname urlstyle\endcsname\relax
  \providecommand{\doi}[1]{doi: #1}\else
  \providecommand{\doi}{doi: \begingroup \urlstyle{rm}\Url}\fi

\bibitem[Ackley et~al.(1985)Ackley, Hinton, and Sejnowski]{ackley1985sampling}
Ackley, D.~H., Hinton, G.~E., and Sejnowski, T.~J.
\newblock A learning algorithm for boltzmann machines.
\newblock \emph{Cognitive Science}, 1985.

\bibitem[Bahdanau et~al.(2017)Bahdanau, Brakel, Xu, Goyal, Lowe, Pineau,
  Courville, and Bengio]{Bahdanau2017actorcritic}
Bahdanau, D., Brakel, P., Xu, K., Goyal, A., Lowe, R., Pineau, J., Courville,
  A., and Bengio, Y.
\newblock \href{https://openreview.net/forum?id=SJDaqqveg}{An Actor-Critic
  Algorithm for Sequence Prediction}.
\newblock In \emph{Proceedings of the $5^{th}$ International Conference on
  Learning Representations (ICLR)}, 2017.

\bibitem[Bengio et~al.(2015)Bengio, Vinyals, Jaitly, and
  Shazeer]{bengio2015scheduledsampling}
Bengio, S., Vinyals, O., Jaitly, N., and Shazeer, N.
\newblock
  \href{https://papers.nips.cc/paper/5956-scheduled-sampling-for-sequence-prediction-with-recurrent-neural-networks}{Scheduled
  Sampling for Sequence Prediction with Recurrent Neural Networks}.
\newblock In \emph{Advances in Neural Information Processing Systems 28
  (NIPS)}, 2015.

\bibitem[Bhattacharyya et~al.(2021)Bhattacharyya, Rooshenas, Naskar, Sun,
  Iyyer, and McCallum]{bhattacharyya2021energybased}
Bhattacharyya, S., Rooshenas, A., Naskar, S., Sun, S., Iyyer, M., and McCallum,
  A.
\newblock Energy-based reranking: Improving neural machine translation using
  energy-based models.
\newblock \emph{arXiv}, 2021.

\bibitem[Browne et~al.(2012)Browne, Powley, Whitehouse, Lucas, Cowling,
  Rohlfshagen, Tavener, Perez~Liebana, Samothrakis, and Colton]{browne2012mcts}
Browne, C., Powley, E., Whitehouse, D., Lucas, S., Cowling, P., Rohlfshagen,
  P., Tavener, S., Perez~Liebana, D., Samothrakis, S., and Colton, S.
\newblock A survey of monte carlo tree search methods.
\newblock \emph{IEEE Transactions on Computational Intelligence and AI in
  Games}, 2012.

\bibitem[Choshen et~al.(2019)Choshen, Fox, Aizenbud, and
  Abend]{choshen2019rlweakness}
Choshen, L., Fox, L., Aizenbud, Z., and Abend, O.
\newblock On the weaknesses of reinforcement learning for neural machine
  translation.
\newblock In \emph{Proceedings of the $8^{th}$ International Conference on
  Learning Representations (ICLR)}, 2019.

\bibitem[Cohen \& Beck(2019)Cohen and Beck]{cohen2019bs}
Cohen, E. and Beck, J.~C.
\newblock Empirical analysis of beam search performance degradation in neural
  sequence models.
\newblock In \emph{Proceedings of the $36^{th}$ International Conference on
  Machine Learning (ICML)}, 2019.

\bibitem[Collobert et~al.(2019)Collobert, Hannun, and
  Synnaeve]{collobert2019fdbs}
Collobert, R., Hannun, A., and Synnaeve, G.
\newblock A fully differentiable beam search decoder.
\newblock In \emph{Proceedings of the $36^{th}$ International Conference on
  Machine Learning (ICML)}, 2019.

\bibitem[Coulom(2006)]{Coulom2006mcts}
Coulom, R.
\newblock Efficient selectivity and backup operators in monte-carlo tree
  search.
\newblock In \emph{Proceedings of the $5^{th}$ International Conference on
  Computers and Games (ICCG)}, 2006.

\bibitem[Denkowski \& Lavie(2014)Denkowski and Lavie]{denkowski2014meteor}
Denkowski, M. and Lavie, A.
\newblock Meteor universal: Language specific translation evaluation for any
  target language.
\newblock In \emph{Proceedings of the EACL 2014 Workshop on Statistical Machine
  Translation}, 2014.

\bibitem[Devlin et~al.(2019)Devlin, Chang, Lee, and Toutanova]{devlin2019bert}
Devlin, J., Chang, M.-W., Lee, K., and Toutanova, K.
\newblock {BERT}: Pre-training of deep bidirectional transformers for language
  understanding.
\newblock In \emph{Proceedings of the 2019 Conference of the North {A}merican
  Chapter of the Association for Computational Linguistics}, 2019.

\bibitem[Edunov et~al.(2018)Edunov, Ott, Auli, Grangier, and
  Ranzato]{edunov2018splosses}
Edunov, S., Ott, M., Auli, M., Grangier, D., and Ranzato, M.
\newblock \href{http://aclweb.org/anthology/N18-1033}{Classical structured
  prediction losses for sequence to sequence learning}.
\newblock In \emph{Proceedings of the 2018 Conference of the North American
  Chapter of the Association for Computational Linguistics (NAACL)}, 2018.

\bibitem[Eikema \& Aziz(2020)Eikema and Aziz]{Eikema2020MBR}
Eikema, B. and Aziz, W.
\newblock Is map decoding all you need? the inadequacy of the mode in neural
  machine translation.
\newblock \emph{ArXiv}, abs/2005.10283, 2020.

\bibitem[Fan et~al.(2018)Fan, Lewis, and Dauphin]{fan2018topksampling}
Fan, A., Lewis, M., and Dauphin, Y.
\newblock Hierarchical neural story generation.
\newblock In \emph{Proceedings of the 56th Annual Meeting of the Association
  for Computational Linguistics (Volume 1: Long Papers)}, 2018.

\bibitem[Gu et~al.(2018)Gu, Bradbury, Xiong, Li, , and Socher]{gu2018nat}
Gu, J., Bradbury, J., Xiong, C., Li, V.~O., , and Socher, R.
\newblock Non-autoregressive neural machine translation.
\newblock In \emph{Proceedings of the $6^{th}$ International Conference on
  Learning Representations (ICLR)}, 2018.

\bibitem[Gumbel(1954)]{gumbel1954max}
Gumbel, E.~J.
\newblock Statistical theory of extreme values and some practical applications:
  a series of lectures, 1954.

\bibitem[Harris et~al.(2020)Harris, Millman, van~der Walt, Gommers, Virtanen,
  Cournapeau, Wieser, Taylor, Berg, Smith, Kern, Picus, Hoyer, van Kerkwijk,
  Brett, Haldane, del R{'{\i}}o, Wiebe, Peterson, G{'{e}}rard-Marchant,
  Sheppard, Reddy, Weckesser, Abbasi, Gohlke, and Oliphant]{harris2020numpy}
Harris, C.~R., Millman, K.~J., van~der Walt, S.~J., Gommers, R., Virtanen, P.,
  Cournapeau, D., Wieser, E., Taylor, J., Berg, S., Smith, N.~J., Kern, R.,
  Picus, M., Hoyer, S., van Kerkwijk, M.~H., Brett, M., Haldane, A., del
  R{'{\i}}o, J.~F., Wiebe, M., Peterson, P., G{'{e}}rard-Marchant, P.,
  Sheppard, K., Reddy, T., Weckesser, W., Abbasi, H., Gohlke, C., and Oliphant,
  T.~E.
\newblock Array programming with {NumPy}.
\newblock \emph{Nature}, September 2020.

\bibitem[He et~al.(2017)He, Lu, Xia, Qin, Wang, and Liu]{he2017decodingvalue}
He, D., Lu, H., Xia, Y., Qin, T., Wang, L., and Liu, T.-Y.
\newblock Decoding with value networks for neural machine translation.
\newblock In \emph{Advances in Neural Information Processing Systems}, 2017.

\bibitem[Holtzman et~al.(2019)Holtzman, Buys, Du, Forbes, and
  Choi]{holtzman2019nucleus}
Holtzman, A., Buys, J., Du, L., Forbes, M., and Choi, Y.
\newblock The curious case of neural text degeneration.
\newblock In \emph{Proceedings of the $7^{th}$ International Conference on
  Learning Representations (ICLR)}, 2019.

\bibitem[Kingma \& Ba(2015)Kingma and Ba]{kingma2014adam}
Kingma, D.~P. and Ba, J.
\newblock Adam: {A} method for stochastic optimization.
\newblock In \emph{3rd International Conference on Learning Representations,
  {ICLR} 2015, San Diego, CA, USA, May 7-9, 2015, Conference Track
  Proceedings}, 2015.

\bibitem[Kocsis \& Szepesv\'ari(2006)Kocsis and Szepesv\'ari]{Kocsis2006mcts}
Kocsis, L. and Szepesv\'ari, C.
\newblock Bandit based monte-carlo planning.
\newblock In \emph{Proceedings of the $15^{th}$ European Conference on Machine
  Learning (ECML)}, 2006.

\bibitem[Kool et~al.(2019)Kool, Van~Hoof, and Welling]{kool2019stochasticbeams}
Kool, W., Van~Hoof, H., and Welling, M.
\newblock Stochastic beams and where to find them: The {G}umbel-top-k trick for
  sampling sequences without replacement.
\newblock In \emph{Proceedings of the $36^{th}$ International Conference on
  Machine Learning (ICML)}, 2019.

\bibitem[Leblond et~al.(2018)Leblond, Alayrac, Osokin, and
  Lacoste-Julien]{leblond2018searnn}
Leblond, R., Alayrac, J.-B., Osokin, A., and Lacoste-Julien, S.
\newblock \href{https://openreview.net/forum?id=HkUR_y-RZ}{\textsc{searnn}:
  training rnns with global-local losses}.
\newblock In \emph{Proceedings of the $6^{th}$ International Conference on
  Learning Representations (ICLR)}, 2018.

\bibitem[Martins et~al.(2017)Martins, Junczys-Dowmunt, Kepler, Astudillo,
  Hokamp, and Grundkiewicz]{martins2017qe}
Martins, A. F.~T., Junczys-Dowmunt, M., Kepler, F.~N., Astudillo, R., Hokamp,
  C., and Grundkiewicz, R.
\newblock Pushing the limits of translation quality estimation.
\newblock \emph{Transactions of the Association for Computational Linguistics},
  2017.

\bibitem[Meister et~al.(2020{\natexlab{a}})Meister, Cotterell, and
  Vieira]{meister-etal-2020-beam}
Meister, C., Cotterell, R., and Vieira, T.
\newblock If beam search is the answer, what was the question?
\newblock In \emph{Proceedings of the 2020 Conference on Empirical Methods in
  Natural Language Processing (EMNLP)}, pp.\  2173--2185. Association for
  Computational Linguistics, November 2020{\natexlab{a}}.

\bibitem[Meister et~al.(2020{\natexlab{b}})Meister, Vieira, and
  Cotterell]{meister2020bestfirst}
Meister, C., Vieira, T., and Cotterell, R.
\newblock Best-first beam search.
\newblock \emph{Transactions of the Association for Computational Linguistics},
  2020{\natexlab{b}}.

\bibitem[Norouzi et~al.(2016)Norouzi, Bengio, Chen, Jaitly, Schuster, Wu, and
  Schuurmans]{Norouzi2016b}
Norouzi, M., Bengio, S., Chen, Z., Jaitly, N., Schuster, M., Wu, Y., and
  Schuurmans, D.
\newblock
  \href{https://papers.nips.cc/paper/6547-reward-augmented-maximum-likelihood-for-neural-structured-prediction}{Reward
  Augmented Maximum Likelihood for Neural Structured Prediction}.
\newblock In \emph{Advances in Neural Information Processing Systems 29
  (NIPS)}, 2016.

\bibitem[Papineni et~al.(2002)Papineni, Roukos, Ward, and
  Zhu]{papineni2002bleu}
Papineni, K., Roukos, S., Ward, T., and Zhu, W.-J.
\newblock \href{https://www.aclweb.org/anthology/P02-1040.pdf}{Bleu: a method
  for automatic evaluation of machine translation}.
\newblock In \emph{Proceedings of the $40^{th}$ Annual Meeting of the
  Association for Computational Linguistics (ACL)}, 2002.

\bibitem[Ranzato et~al.(2016)Ranzato, Chopra, Auli, and Zaremba]{Ranzato2016b}
Ranzato, M., Chopra, S., Auli, M., and Zaremba, W.
\newblock \href{https://arxiv.org/abs/1511.06732}{Sequence Level Training with
  Recurrent Neural Networks}.
\newblock In \emph{Proceedings of the $5^{th}$ International Conference on
  Learning Representations (ICLR)}, 2016.

\bibitem[Reddy(1977)]{reddy}
Reddy, R.
\newblock \emph{Speech understanding systems: A summary of results of the
  five-year research effort}.
\newblock Carnegie Mellon University, 1977.

\bibitem[{Ren} et~al.(2017){Ren}, {Wang}, {Zhang}, {Lv}, and
  {Li}]{ren2017captioning}
{Ren}, Z., {Wang}, X., {Zhang}, N., {Lv}, X., and {Li}, L.
\newblock Deep reinforcement learning-based image captioning with embedding
  reward.
\newblock In \emph{2017 IEEE Conference on Computer Vision and Pattern
  Recognition (CVPR)}, 2017.

\bibitem[Schrittwieser et~al.(2020)Schrittwieser, Antonoglou, Hubert, Simonyan,
  Sifre, Schmitt, Guez, Lockhart, Hassabis, Graepel, Lillicrap, and
  Silver]{schrittwieser2020muzero}
Schrittwieser, J., Antonoglou, I., Hubert, T., Simonyan, K., Sifre, L.,
  Schmitt, S., Guez, A., Lockhart, E., Hassabis, D., Graepel, T., Lillicrap,
  T., and Silver, D.
\newblock Mastering atari, go, chess and shogi by planning with a learned
  model.
\newblock \emph{Nature}, 2020.

\bibitem[Shazeer(2019)]{shazeer2019singlekv}
Shazeer, N.
\newblock Fast transformer decoding: One write-head is all you need.
\newblock \emph{arXiv}, abs/1911.02150, 2019.

\bibitem[Shen et~al.(2016)Shen, Cheng, He, He, Wu, Sun, and Liu]{Shen2016mrt}
Shen, S., Cheng, Y., He, Z., He, W., Wu, H., Sun, M., and Liu, Y.
\newblock \href{http://www.aclweb.org/anthology/P16-1159}{Minimum Risk Training
  for Neural Machine Translation}.
\newblock In \emph{Proceedings of the $53^{rd}$ Annual Meeting of the
  Association for Computational Linguistics (ACL)}, 2016.

\bibitem[Silver et~al.(2017)Silver, {Schrittwieser Julian}, {Simonyan Karen},
  {Antonoglou Ioannis}, {Huang Aja}, {Guez Arthur}, {Hubert Thomas}, {Baker
  Lucas}, {Lai Matthew}, {Bolton Adrian}, {Chen Yutian}, {Lillicrap Timothy},
  {Hui Fan}, {Sifre Laurent}, {van den Driessche George}, {Graepel Thore}, and
  {Hassabis Demis}]{silver2017alphago}
Silver, D., {Schrittwieser Julian}, {Simonyan Karen}, {Antonoglou Ioannis},
  {Huang Aja}, {Guez Arthur}, {Hubert Thomas}, {Baker Lucas}, {Lai Matthew},
  {Bolton Adrian}, {Chen Yutian}, {Lillicrap Timothy}, {Hui Fan}, {Sifre
  Laurent}, {van den Driessche George}, {Graepel Thore}, and {Hassabis Demis}.
\newblock \href{http://dx.doi.org/10.1038/nature24270}{Mastering the game of Go
  without human knowledge}.
\newblock \emph{Nature}, 2017.

\bibitem[Specia et~al.(2018)Specia, Scarton, and Paetzold]{specia2018qe}
Specia, L., Scarton, C., and Paetzold, G.~H.
\newblock Quality estimation for machine translation.
\newblock \emph{Synthesis Lectures on Human Language Technologies}, 2018.

\bibitem[Stahlberg \& Byrne(2019)Stahlberg and Byrne]{StahlbergByrne2019nmt}
Stahlberg, F. and Byrne, B.
\newblock On {NMT} search errors and model errors: Cat got your tongue?
\newblock In \emph{Proceedings of the 2019 Conference on Empirical Methods in
  Natural Language Processing and the 9th International Joint Conference on
  Natural Language Processing (EMNLP-IJCNLP)}, 2019.

\bibitem[Vaswani et~al.(2017)Vaswani, Shazeer, Parmar, Uszkoreit, Jones, Gomez,
  Kaiser, and Polosukhin]{vaswani2017transformer}
Vaswani, A., Shazeer, N., Parmar, N., Uszkoreit, J., Jones, L., Gomez, A.~N.,
  Kaiser, L.~u., and Polosukhin, I.
\newblock Attention is all you need.
\newblock In \emph{Advances in Neural Information Processing Systems}, 2017.

\bibitem[Wang et~al.(2020)Wang, Kuo, and Katz]{Wang2020bslookahead}
Wang, Y.-S., Kuo, Y.-L., and Katz, B.
\newblock Investigating the decoders of maximum likelihood sequence models: A
  look-ahead approach.
\newblock \emph{ArXiv}, abs/2003.03716, 2020.

\bibitem[Welleck et~al.(2019)Welleck, Brantley, Daum{\'{e}}, and
  Cho]{welleck2019outoforder}
Welleck, S., Brantley, K., Daum{\'{e}}, III, H., and Cho, K.
\newblock Non-monotonic sequential text generation.
\newblock In \emph{Proceedings of the $36^{th}$ International Conference on
  Machine Learning (ICML)}, 2019.

\bibitem[Welleck et~al.(2020)Welleck, Kulikov, Kim, Pang, and
  Cho]{welleck2020consistency}
Welleck, S., Kulikov, I., Kim, J., Pang, R.~Y., and Cho, K.
\newblock Consistency of a recurrent language model with respect to incomplete
  decoding.
\newblock In \emph{Proceedings of the 2020 Conference on Empirical Methods in
  Natural Language Processing (EMNLP)}, 2020.

\bibitem[Wu et~al.(2016)Wu, Schuster, Chen, Le, Norouzi, Macherey, Krikun, Cao,
  Gao, Macherey, Klingner, Shah, Johnson, Liu, Kaiser, Gouws, Kato, Kudo,
  Kazawa, Stevens, Kurian, Patil, Wang, Young, Smith, Riesa, Rudnick, Vinyals,
  Corrado, Hughes, and Dean]{wu2016bridging}
Wu, Y., Schuster, M., Chen, Z., Le, Q.~V., Norouzi, M., Macherey, W., Krikun,
  M., Cao, Y., Gao, Q., Macherey, K., Klingner, J., Shah, A., Johnson, M., Liu,
  X., Kaiser, L., Gouws, S., Kato, Y., Kudo, T., Kazawa, H., Stevens, K.,
  Kurian, G., Patil, N., Wang, W., Young, C., Smith, J., Riesa, J., Rudnick,
  A., Vinyals, O., Corrado, G., Hughes, M., and Dean, J.
\newblock Google's neural machine translation system: Bridging the gap between
  human and machine translation.
\newblock \emph{arXiv}, 2016.
\newblock URL \url{http://arxiv.org/abs/1609.08144}.

\bibitem[Yu et~al.(2020)Yu, Sartran, Huang, Stokowiec, Donato, Srinivasan,
  Andreev, Ling, Mokra, Dal~Lago, Doron, Young, Blunsom, and
  Dyer]{yu2020mctswmt}
Yu, L., Sartran, L., Huang, P.-S., Stokowiec, W., Donato, D., Srinivasan, S.,
  Andreev, A., Ling, W., Mokra, S., Dal~Lago, A., Doron, Y., Young, S.,
  Blunsom, P., and Dyer, C.
\newblock The {D}eep{M}ind {C}hinese{--}{E}nglish document translation system
  at {WMT}2020.
\newblock In \emph{Proceedings of the Fifth Conference on Machine Translation},
  2020.

\bibitem[Zhang et~al.(2019)Zhang, Kishore, Wu, Weinberger, and
  Artzi]{zhang2019bertscore}
Zhang, T., Kishore, V., Wu, F., Weinberger, K.~Q., and Artzi, Y.
\newblock {BERTS}core: Evaluating text generation with {BERT}.
\newblock In \emph{Proceedings of the $7^{th}$ International Conference on
  Learning Representations (ICLR)}, 2019.

\end{thebibliography}
\bibliographystyle{icml2021}

\clearpage

\appendix
\section*{Outline}

Appendix~\ref{apx:hybrid} provides details about the network architectures and the optimisation hyper-parameters used in our work.
We also describe the hybrid architecture that is used to improve the training of value networks in the hard case of \emph{privileged} metrics.
In Appendix~\ref{apx:expdet}, we lay out the details of the tuning of the different decoding mechanisms used throughout the paper.
In Appendix~\ref{apx:scaling}, we give results about how performance of our decoding algorithms scale with more computational budget.
In Appendix~\ref{apx:supdistill}, we discuss the trade offs between learning a joint policy and value network on sampled (or distilled) trajectories, versus training a separate value network to predict the value of a policy network trained on supervised trajectories.
In Appendix~\ref{apx:qual}, we give a few examples of MCTS exploratory trees.
Finally, in Appendix~\ref{apx:code} we provide a simple implementation of a batched version of MCTS in plain Numpy.

\section{Network architectures and training}\label{apx:hybrid}
In this section, we detail our basic dual-headed architecture, our training regimen and our optimisation hyper-parameters.
We also describe our hybrid architecture (which we use for privileged metrics) in more depth.

\paragraph{Dual-head transformer architecture.}
We start from the original transformer encoder-decoder architecture~\citep{vaswani2017transformer}, with a few modifications.
Both encoder and decoder have $\mathrm{num\_layers} = 6$ attention layers.
The hidden size is 512.
The embedding vocabulary size is just short of 32000 tokens for both language pairs.
The unroll length of our models is 128.
We use ``normal GPT-2"-style initialisers, i.e. initial values are sampled from a Gaussian distribution with mean 0 and standard deviation $\frac{0.02}{\sqrt{\mathrm{num\_layers}}}$.
The one exception to this rule is for embeddings, where we use truncated normal initialisers with standard deviation 1.0.

On top of the decoder, we add two ``heads".
The first one is the policy head. 
It consists in a linear projection from the hidden dimensionality to the vocabulary size, followed by a softmax operator to output a distribution over the whole vocabulary.
The second head is the value head: a linear projection from the hidden dimensionality to the amount of value buckets we define ($|\mathcal{B}| = 500$ in our experiments), followed by a softmax operator.
We compute the value loss as the cross-entropy between the softmax distribution $(v_i)_{i \in \mathcal{B}}$ and a one-hot encoding of the target value of the same dimension.
To output the value, we compute the sum of the softmax distribution multiplied by the average value in each bucket $\sum_{i \in \mathcal{B}} v_i \bar b_i$.

Compared to the original architecture, we apply several changes related to inference speed.
First, instead of 8 attention heads we use 16.
Second, the dimensionality of the keys and values is 128, compared to 64 in the original architecture, thus avoiding a costly padding operation on our hardware accelerators, TPUv3.
Finally, we use multi-query attention~\citep{shazeer2019singlekv}, only computing a single set of keys and values per attention block and sharing them across all attention heads.
This reduces the memory footprint of the keys and values by a factor of the number of attention heads (16 here), considerably decreasing the time spent reading and writing from memory, which ultimately results in a near-linear inference speedup with respect to the number of attention heads.
As this alternative attention mechanism requires less trainable weights than the more conventional one, we reallocate some of those in the feedforward layer of the attention blocks by using a bigger internal hidden dimensionality of 3072 instead of 2048.

\paragraph{Optimisation.}
We use the Adam~\citep{kingma2014adam} optimiser with learning rate 0.001, and the following hyper-parameters: $b_1 = 0.9; b_2 = 0.98; \epsilon = 1e^{-9}$.
Our batch size is 4096, and we train for 100000 steps for the ENDE dataset and 300000 steps for the larger ENFR dataset.
As regularisation, we use dropout with a weight 0.1, but no weight decay.
We also use label smoothing with hyper-parameter 0.1 (although we see little impact when removing it).

The last difference with the original transformer encoder-decoder is where we place the layer norm operator. 
We put it at the beginning of the attention and the feed-forward layers, rather than at the end, which allows for fully-residual layers.

\paragraph{Hybrid architecture for privileged metrics.}
In Section~\ref{value}, we show that learning good value functions on privileged metrics such as BLEU and BERTScore is very difficult.
This is mainly due to the fact that our value networks are lacking access to the ground truth targets which are required for precise score computation.
Our ablation study show that if we remove this difficulty by allowing the value model to ``cheat'' by using the ground truth targets rather than the source sentences as encoder inputs, we obtain much more precise values.
Downstream results using MCTS with such a ``cheating'' value show very large BLEU improvements.
Unfortunately, in practice one cannot rely on such a trick when decoding.

Our idea is to try to leverage ``cheating'' information indirectly at training time to shape the representation of a regular (i.e. non-cheating) value network.
Another way to look at it is that we try to distill the knowledge of the cheating value model into the regular one.

\begin{figure}
    \centering
    \includegraphics[width=\linewidth]{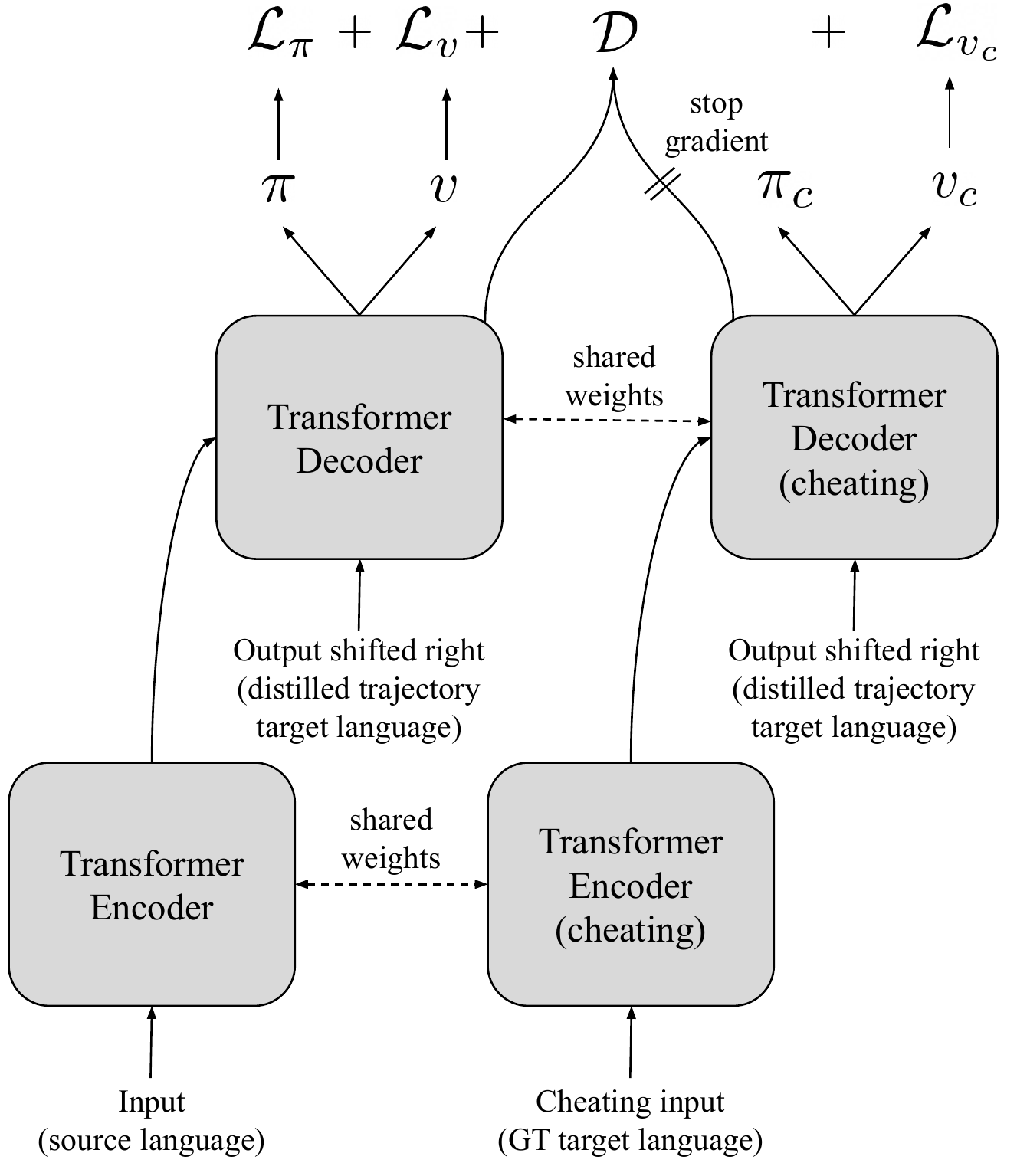}
    \caption{
    \small
    \textbf{Hybrid architecture.}
    Since learning a good value network is hard when dealing with \emph{privileged} metrics, we propose a training mechanism to distill the information from a \emph{cheating} network that has access to the full \emph{privileged} information in order to predict the value score.
    In details, we have two encoder/decoder networks that share their weights. 
    The first network (\emph{left}) is a regular network that takes as input the input sentence in the source language and is trained to output (i) a policy predicting the  words of the output sentence in the target language and (ii) a value score for that output via a policy and a value loss $\mathcal{L}_\pi$ and $\mathcal{L}_v$, respectively. 
    The second network (\emph{right}), dubbed cheating, is given as input the ground truth sentence in the target language and is trained to output the value score of the output sentence in the target language against that ground truth via the loss $\mathcal{L}_{v_c}$.
    This simplifies its task considerably, as it has direct access to the \emph{privileged} information to compute the value.
    A distillation loss $\mathcal{D}$ is added to transfer knowledge from the cheating model to the regular network.
    Results in Table~\ref{table:hybrid} shows that such an approach yields significant improvements by easing the training of the value network.
    }
    \label{fig:hybrid}
\end{figure}

To achieve this, we propose a new training regimen, as detailed in Figure~\ref{fig:hybrid}.
The basic idea is to compute the final layer of a cheating value model, and to use it as an auxiliary target for the final layer of a regular value model, in addition to its normal value loss.
We thus have two pathways. 
On the left, the regular value model encoder receives the source sentence as input (which is available at test time).
On the right, the cheating value model encoder receives the ground truth target sentence as input (which is \textbf{not} available at test time).
For both pathways, the decoder's input is a sample sentence.
Crucially, both pathways rely on the same transformer encoder-decoder: they share all weights, the only difference is in their inputs.

To train such a model, we use four losses.
First, we apply the regular policy $\mathcal{L}_{\pi}$ and value loss $\mathcal{L}_v$ on the regular pathway.
Second, we apply a value loss $\mathcal{L}_{v_c}$ on the cheating pathway.
Finally, we add an $\ell_2$ loss $\mathcal{D}$ between the final layer of both pathways, with a stop gradient for the cheating one -- so that its representation is not directly affected by $\mathcal{D}$.

We do not add a policy loss on the cheating pathway.
This seems natural, as such a loss would only encourage the model to reproduce its inputs exactly, effectively pushing it towards the identity function.

\begin{table}[]
\begin{tabular*}{\textwidth / 2 - \tabcolsep}{@{\extracolsep{15pt}}lccc@{}}
\toprule
Architecture & Normal & Cheating & Hybrid \\ \midrule
Greedy       & 26.11  & --       & 25.99  \\
MCTS         & 26.40  & \gray{38.52}    & 27.47  \\ \midrule
Greedy       & 87.92  & --       & 87.88  \\
MCTS         & 88.01  & \gray{90.70}    & 88.48  \\ \bottomrule
\end{tabular*}
\caption{Greedy vs MCTS (50 simulations) performance on the ENDE dataset for BLEU (top rows) and BERTScore (bottom rows). Using the normal architecture, improvements are very small. Using the hybrid architecture yields more significant improvements. The middle column contains (greyed-out) results when using a cheating model which takes ground truth targets as inputs. Improvements are enormous in this prohibited setting, which is unsurprising at the value function receives optimal output as its own inputs.}
\label{table:hybrid}
\end{table}

Using such a hybrid architecture yields performance improvements when using the value model with MCTS, as shown in Table~\ref{table:hybrid}.

We find that to obtain best performance, three things need to be combined: (i) sharing weights across both pathways, (ii) the distillation $\ell_2$ loss and (iii) the cheating value loss.
Each loss is added with a linear weight.
Proper tuning of these weights is important.
We find that using weights 1.0 for $\mathcal{L}_{\pi}$ and $\mathcal{L}_{v_c}$, as well as 0.1 for $\mathcal{L}_v$ and $\mathcal{D}$ leads to best performance.

\section{Experimental details and ablations}\label{apx:expdet}
We detail how we tuned each algorithm for best performance in this section.
We used the dev datasets to determine which options were best.
Wherever we report numbers, we compute those on the test set for comparison purposes, but the experiments were run \emph{after} dev set selection.
In practice, we find a very good correlation between observations on the dev and on the test sets (although absolute values were lower on the dev set, rankings remained mostly unchanged).
Unless otherwise indicated, our findings hold for both ENDE and ENFR datasets, and for all 3 metrics we consider (BLEU, BERTScore, Multilingual BERTScore).

\paragraph{Beam Search.}
As this method is known for degrading with large beam sizes, we add a length normalisation term, as advocated by~\citet{wu2016bridging}.
The resulting score for a candidate $y_1 .. y_t$ is thus:
$bs(y_1 .. y_t) = (\frac{6}{t + 5})^\theta \log\big(\pi(y_1 .. y_t)\big)$

We tuned three hyper-parameters for beam search:
\begin{itemize}
    \item the beam size: we tried 2, 4, 6, 8, 10 and 20. We find that the best performance is attained at 6, plateaus until 10 and starts slowly decreasing by 20 (see Table~\ref{table:beamsize}).
    \item the logits temperature: we tried 0.6, 0.8, 1.0, 1.2, 1.4. 1.0 performs best by a wide margin. Low values degrade to greedy search performance while high values yield non-sensical sentences.
    \item the normalisation temperature parameter $\theta$: we tried 0.4, 0.6, 0.8 and 1.0. We find $\theta = 0.6$ performs best, as in the original paper ($\theta = 0.4$ is on par but slightly worse, and performance degrades as soon as $\theta \geq 0.8$).
\end{itemize}

\begin{table}[]
\begin{tabular*}{\textwidth / 2 - \tabcolsep}{@{\extracolsep{55pt}}rcc@{}}
\toprule
   & ENDE  & ENFR  \\ \midrule
2  & 27.22 & 38.88 \\
4  & 27.75 & 39.24 \\
6  & \textbf{28.03} & \textbf{39.37} \\
8  & 28.03 & 39.29 \\
10 & 28.01 & 39.29 \\
20 & 28.02 & 39.21 \\ \bottomrule
\end{tabular*}
\caption{BLEU results as a function of beam size for plain beam search.}
\label{table:beamsize}
\end{table}

\paragraph{Value-guided beam search.}
The score for a candidate $y_1 .. y_t$ for this method is:
\begin{equation*}
    bs(y_1 .. y_t) = \frac{\alpha}{t} \log\big(\pi(y_1 .. y_t)\big) + (1 - \alpha) v(y_1 .. y_t).
\end{equation*}

The hyper-parameters to tune are slightly different than those of plain beam search. 
As we do not see performance decrease with larger beam size, we no longer need to find the optimal one.
Which one to use is largely dependent on how much computation one can afford to use.
Performance as a function of this quantity are reported in Section~\ref{apx:scaling}.

Further, because of the additional value term, we have a new linear combination weight $\alpha$ to tune.
Here are the hyper-parameter ranges we consider:
\begin{itemize}
    \item the logits temperature: as for plain beam search, we tried 0.6, 0.8, 1.0, 1.2, 1.4. We find similar results: 1.0 is the best-performing temperature by a significant margin; the reliance on the value mitigates the effect for unprivileged metrics (where the value is a good approximation).
    \item the value normalisation: we tried using the logarithm of the value instead of the value itself, with appropriate scaling: $\log(\frac{v(y_1..y_t) - d}{D - d})$.
    We find that by carefully tuning the minimum and maximum bounds $d$ and $D$ we can match the performance we obtain with the plain value, but not outperform it. As a result we opt for the simpler formulation.
    \item the linear combination weights $\alpha$: we swept between 0 and 1 by 0.1 increments (with an additional measure at 0.95). For privileged metrics, we find $\alpha = 0.5$ to perform best (on both datasets). For Multilingual BERTScore on the other hand, much larger values are required to achieve best performance: $\alpha=0.9$ for ENDE and $\alpha = 0.95$ for ENFR. Our hypothesis is that the quality of our value function is much higher for this last~\emph{unprivileged} metric, which allows us to lean more heavily on its guidance. See Table~\ref{table:vgbsalpha} for illustrative results.
\end{itemize}

\begin{table}[]
\begin{tabular*}{\textwidth / 2 - \tabcolsep}{@{\extracolsep{35pt}}lcc@{}}
\toprule
      & \multicolumn{2}{c}{ENFR} \\ \cmidrule(l){2-3} 
Score & BERTScore  & MLBERTScore \\ \midrule
0.0   & 90.74      & 84.41       \\
0.1   & 90.77      & 84.42       \\
0.2   & 90.80      & 84.44       \\
0.3   & 90.80      & 84.51       \\
0.4   & 90.83      & 84.62       \\
0.5   & \textbf{90.84}      & 84.77       \\
0.6   & 90.71      & 84.90       \\
0.7   & 89.91      & 85.10       \\
0.8   & 87.17      & 85.44       \\
0.9   & 82.25      & 85.44       \\
0.95  & 79.47      & \textbf{85.55}       \\
1.0   & 76.22      & 79.30       \\ \bottomrule
\end{tabular*}
\caption{VGBS performance on the ENFR dataset as a function of the linear weight $\alpha$.}
\label{table:vgbsalpha}
\end{table}

\paragraph{MCTS variants.}
MCTS is a complex algorithm with a large number of hyper-parameters.
We found that three main aspects are important for performance: making sure the policy and the value terms are well-balanced in the UCT formula, picking the best value aggregation mechanism during the backup phase, and selecting the best acting criteria (once the tree is finished).

To ensure balance in the UCT formula, we tuned two things:
\begin{itemize}
    \item we optimised for the logits temperature $\tau$ and the multiplicative constant $c_{\mathrm{puct}}$ \emph{jointly}. We tried temperatures 0.9, 1.1 and 1.3, in conjunction with $c_{\mathrm{puct}}$ in 1.0, 2.0, 3.0, 4.0, 6.0, 8.0. For privileged metrics, the pair $(\tau = 0.9; c_{\mathrm{puct}} = 3.0)$ performed best across both datasets and both metrics; while for Multilingual BERTScore the best performer was $(\tau = 1.1; c_{\mathrm{puct}} = 8.0)$ across both datasets. Note that both larger temperature and larger $c_{\mathrm{puct}}$ encourage exploration in the UCT formula, thus reducing the relative weight of the policy in favour of the value; that we can use larger scalars for unprivileged metrics is yet another indication that the associated value functions are more trustworthy.
    \item as we detailed in the main text, we rescale the values dynamically during the tree construction so that all the values encountered until the current step are more evenly distributed in the $[0, 1]$ interval by mapping the minimum value to 0 and the maximum value to 1.
\end{itemize}

We tested two value aggregation operators: running average and maximum.
We also tried two action selection mechanisms: picking among the root's children nodes the one with maximum visit count, or the one with maximum aggregated value.

\begin{table}[]
\begin{tabular*}{\textwidth / 2 - \tabcolsep}{@{\extracolsep{5pt}}lcccc@{}}
\toprule
Action selection      & \multicolumn{2}{c}{argmax(vc)} & \multicolumn{2}{c}{argmax(v)} \\ \cmidrule(l){2-3} \cmidrule(l){4-5} 
Value aggregation       &  avg.   & max.   & avg.    &  max.  \\ \midrule
BLEU                   &       \textbf{27.47}        &       26.82      &       22.48        &     25.52        \\
MLBERTScore &        83.76       &      83.81       &      84.08         & \textbf{84.97}       \\ \bottomrule
\end{tabular*}
\caption{BLEU and Multilingual BERTScore performance when using different value aggregation mechanisms and action selection rules. We observe that on the unprivileged metric, the best option rely more heavily on the value function; contrary to what we see for the privileged metric.}
\label{table:maxmax}
\end{table}
Our observation once again underline the contrast between privileged and unprivileged metrics, as is illustrated in Table~\ref{table:maxmax}.
For the former, the best choice is to use the running average as value aggregation operator during the backup phase, and to pick the root child with maximum visit counts.
Conversely, for Multilingual BERTScore we found that we obtained best performance with the maximum aggregation operator and by picking the root child with maximum aggregated value.
We thus see that for our unprivileged metric we can rely on the value function aggressively; while for privileged metrics we need to limit our exposure to it.

\paragraph{Sampling + Ranking variants.}
For these algorithms, we really only have a single hyper-parameter to tune: the policy temperature $\tau$.
Small temperatures lead to little diversity across different samples, but ensure that samples are highly ranked by the model and hence are syntactically correct.
On the other hand, large temperature encourage diversity, at the price of correctness.
We sweep over the [0.15, 0.95] interval by 0.1 increments, and report results in Table~\ref{table:s+r}.
We find that for the score-based S+R, balancing diversity with correctness means we have to use $\tau = 0.75$.

The story is more nuanced for S+RV.
As this variant relies on the value function rather than the score to rerank samples, we can use it even for privileged metrics.
We observe that for these, the optimal temperature is much smaller ($\tau = 0.25$), which in effect means that the algorithm relies more heavily on the policy, compared to the value.
The reason why is once again that the value for this type of metrics is of lower quality.

In contrast, the optimal temperature for our unprivileged metric is $\tau = 0.75$, similar to what we find for the score-based S+R.

\begin{table}[]
\begin{tabular*}{\textwidth / 2 - \tabcolsep}{@{\extracolsep{8pt}}lcccc@{}}
\toprule
     & S+R                  & \multicolumn{3}{c}{S+RV}                                           \\ \cmidrule(l){2-2} \cmidrule(l){3-5}
     & MLBERT                & BLEU                 & BERT                   & MLBERT                 \\ \midrule
0.15 &      83.97                &   26.17                   &      88.31                &       83.69               \\
0.25 &      84.35                &   \textbf{26.23}                   &      \textbf{88.32}                &         84.02             \\
0.35 &      84.60                &   25.85                   &      88.31                &       84.22               \\
0.45 &      84.82                &   25.47                   &      88.29                &       84.37               \\
0.55 &      84.91                &   24.96                   &      88.14                &       84.44               \\
0.65 &      85.02                &   24.28                   &      88.04                &       84.48               \\
0.75 &      \textbf{85.11}                &   23.79                   &      87.91                &       \textbf{84.48}               \\
0.85 &      85.07                &   22.77                   &      87.69                &       84.21               \\
0.95 &      84.95                &   21.32                   &      87.23                &       83.66
\end{tabular*}
\caption{Sampling + ranking performance as a function of policy logits temperature, for both score and value-based variants.}
\label{table:s+r}
\end{table}

\section{Scaling search with computational budget}\label{apx:scaling}

\begin{table*}[]
\resizebox{1.0\linewidth}{!}{
\footnotesize
\begin{tabular}{@{}rccccccccccccc@{}}
\toprule
\multicolumn{1}{l}{} & \multicolumn{4}{c}{BLEU}      & \multicolumn{4}{c}{BERTScore} & \multicolumn{5}{c}{MLBERTScore}       \\ \cmidrule(l){2-5}\cmidrule(l){6-9}\cmidrule(l){10-14}
\multicolumn{1}{l}{} & BS    & VGBS  & S+RV  & MCTS  & BS    & VGBS  & S+RV  & MCTS  & BS    & VGBS  & S+R   & S+RV  & MCTS  \\ \midrule
1                    & 38.16 & 38.16 & 28.54 & 38.16 & 90.52 & 90.52 & 87.98 & 90.52 & 84.22 & 84.22 & 82.94 & 82.97 & 84.22 \\
10                   & \textbf{38.95} & 39.30 & 34.46 & 38.96 & 90.71 & 90.83 & 90.23 & 90.81 & 84.36 & 85.38 & 85.35 & 85.07 & 85.76 \\
25                   & 38.81 & 39.25 & 35.01 & \textbf{39.12} & \textbf{90.76} & \textbf{90.87} & \textbf{90.40} & \textbf{90.85} & 84.42 & 85.69 & 85.82 & 85.44 & 86.10 \\
50                   & 38.67 & 39.33 & 35.19 & 38.95 & 90.76 & 90.85 & 90.37 & 90.80 & 84.46 & 85.80 & 86.12 & 85.68 & 86.31 \\
75                   & 38.62 & 39.41 & \textbf{35.56} & 38.89 & 90.76 & 90.86 & 90.31 & 90.74 & 84.48 & 85.75 & 86.33 & 85.86 & 86.44 \\
100                  & 38.55 & \textbf{39.42} & 35.13 & 38.84 & 90.76 & 90.86 & 90.37 & 90.67 & 84.48 & 85.78 & 86.44 & 85.95 & 86.51 \\
200                  & 38.54 & 39.39 & 35.31 & 38.30 & 90.76 & 90.84 & 90.26 & 90.33 & 84.48 & 85.77 & 86.76 & 86.22 & 86.62 \\
300                  & 38.45 & 39.36 & 35.00 & 37.70 & 90.74 & 90.83 & 90.16 & 89.83 & 84.49 & \textbf{85.79} & \textbf{86.90} & \textbf{86.33} & \textbf{86.66} \\ \bottomrule
\end{tabular}}
\caption{Comparison of how decoding algorithms scale with computational budget, on the ENFR dataset, for both privileged and unprivileged metrics.}
\label{table:apxscaling}
\end{table*}

We present more detailed scaling results in this section.
These confirm our main observation: the higher the quality of the metric signal we use, the better the method scales with additional computation.

We see for instance that on privileged metrics, where value functions are hard to train, the performance of value-based methods reaches its peak quickly and start degrading.
Comparatively, on unprivileged metrics, value-based methods keep improving with more inferences, eventually plateauing.
Finally, score-based methods do not even plateau.
As a result, S+R ends up outperforming MCTS after 200 simulations per token, although MCTS remains the best performer under 100 simulations.
This motivates investigating a variant of MCTS which is allowed to use the score on completed sentences (our current algorithm is purely value-based).

Beam search, which does not use the metric at all, behaves thus more similarly across all metrics, quickly reaching its peak performance and then plateauing.
The length penalty is crucial to prevent performance degradation.

Another interesting observation is that S+RV, the value-based alternative of S+R, performs worse than VGBS or MCTS.
It appears that the crucial ingredient to S+R good performance is direct access to the score, rather than its simple search mechanism.

Finally, we note that for VGBS, each token costs $k + k^2$ inferences ($k$ to compute the policy for every beam, $k^2$ to compute the value for the $k^2$ possible follow-up tokens).
As a result, we use the smallest $k$ such that $k + k^2 \geq n$ when allowing $n$ simulations for other decoding algorithms.

\section{Supervised policy vs Distilled policy} \label{apx:supdistill}
As we study decoding mechanisms on privileged metrics based on the ground truth, we cannot train our value networks on the initial supervised dataset (where the value target would be 1 for all items, as the ground truth targets are considered optimal).
As a result we go through an intermediate step, first training a supervised policy model, and then replacing ground truth targets by a greedy sample from the said policy to create a new distillation dataset.

We observe is the performance of the policy models trained on the distillation datasets is slightly lower than that of their plainly supervised counterparts, as illustrated in Table~\ref{table:supdis}.
The effect is larger for the bigger dataset, ENFR.

This prompts another line of investigation: what happens if at decoding time we use a value model (trained on a distillation dataset) together with a policy model trained on a supervised dataset?
We present BLEU results for MCTS in Table~\ref{table:supvalue}.
On the larger dataset, we see that MCTS decoding outperforms any other type of decoding.
Interestingly, we obtain an improvement of more than 1 BLEU point over the supervised baseline when using a Multilingual BERTScore value function; and we obtain this result with a relatively low amount of simulations (25).
Unfortunately adding more computational budget does not help, as the decoding is targeting a different metric than BLEU.
But with a low enough amount of simulations, we see that trying to optimise our unprivileged metric yields benefits.

While this result is close to the state of the art for such a small policy model, the comparison is not fair as the approach requires double the amount of parameters (since the value net is another network).
This could be alleviated by training a supervised policy, fixing its weights, and adding a lightweight value head on top in a second training step.
We leave this for future work.

\begin{table}[]
\begin{tabular*}{\textwidth / 2 - \tabcolsep}{@{\extracolsep{25pt}}lcc@{}}
\toprule
Policy model & Supervised & Distilled \\ \midrule
ENDE         & \textbf{25.99}      & 25.95     \\
ENFR         & \textbf{38.70}      & 38.16     \\ \bottomrule
\end{tabular*}
\caption{Greedy decoding performance for plainly supervised policy model and dual-headed distilled ones.}
\label{table:supdis}
\end{table}

\begin{table}[]
\begin{tabular*}{\textwidth / 2 - \tabcolsep}{@{\extracolsep{5pt}}lcccc@{}}
\toprule
     & \multirow{2}{*}{Beam search} & \multicolumn{3}{c}{Value type}  \\ \cmidrule(l){3-5} 
     &                              & BLEU  & BERT & MLBERT \\ \midrule
ENDE & 27.75                        & 27.33 & 27.42     & 27.17       \\
ENFR & 39.24                        & 39.67 & 39.46     & \textbf{40.31}       \\ \bottomrule
\end{tabular*}
\caption{MCTS (25 simulations) BLEU performance, using a supervised policy model and a distinct value model.}
\label{table:supvalue}
\end{table}

\onecolumn
\section{MCTS examples} \label{apx:qual}
\begin{figure}[H]
\centering
\includegraphics[scale=0.36]{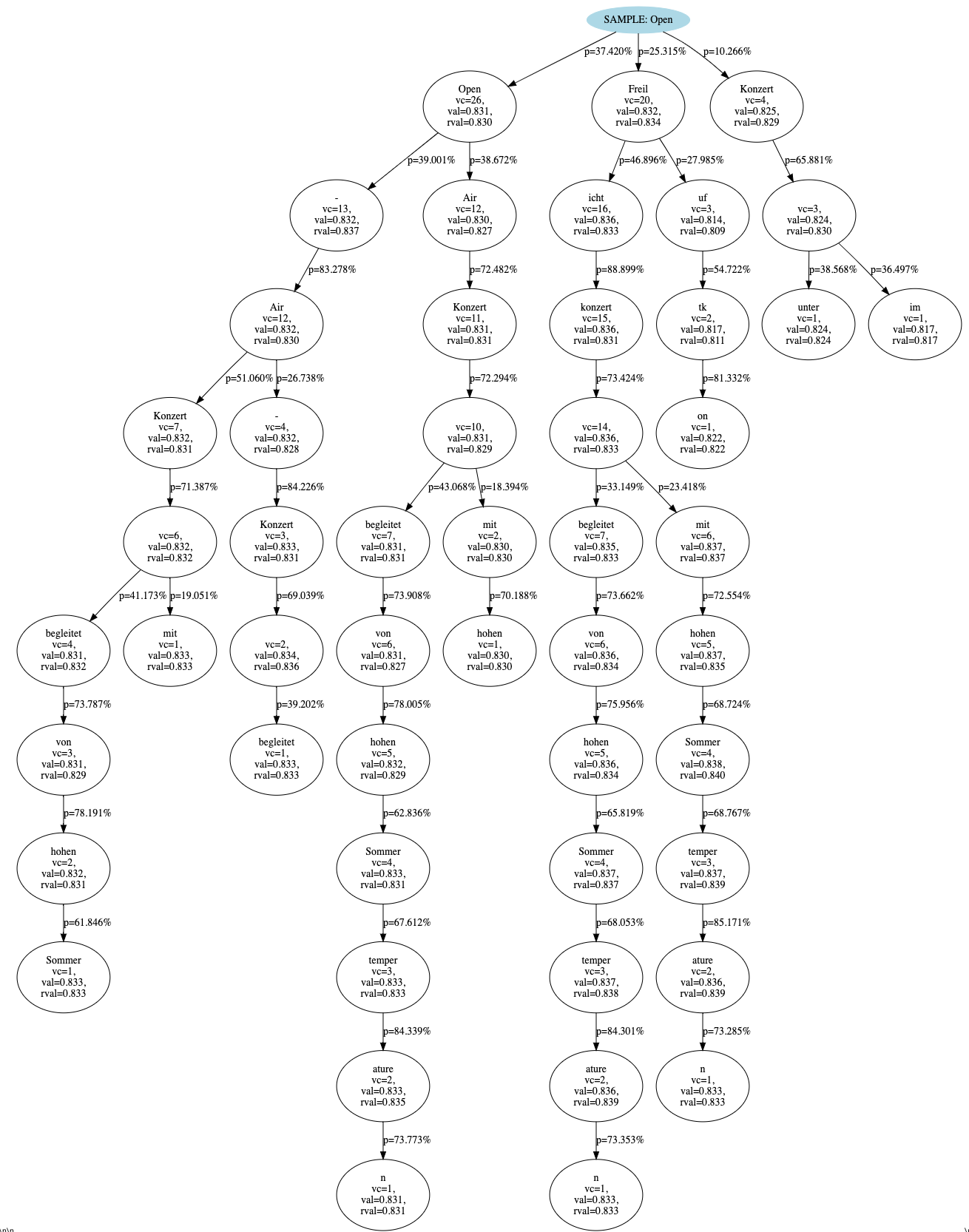}
\caption{MCTS tree when translating ``Open-air concert accompanied by high summer temperatures" from English to German, first step.}
\end{figure}

\begin{figure}[H]
\centering
\includegraphics[scale=0.4]{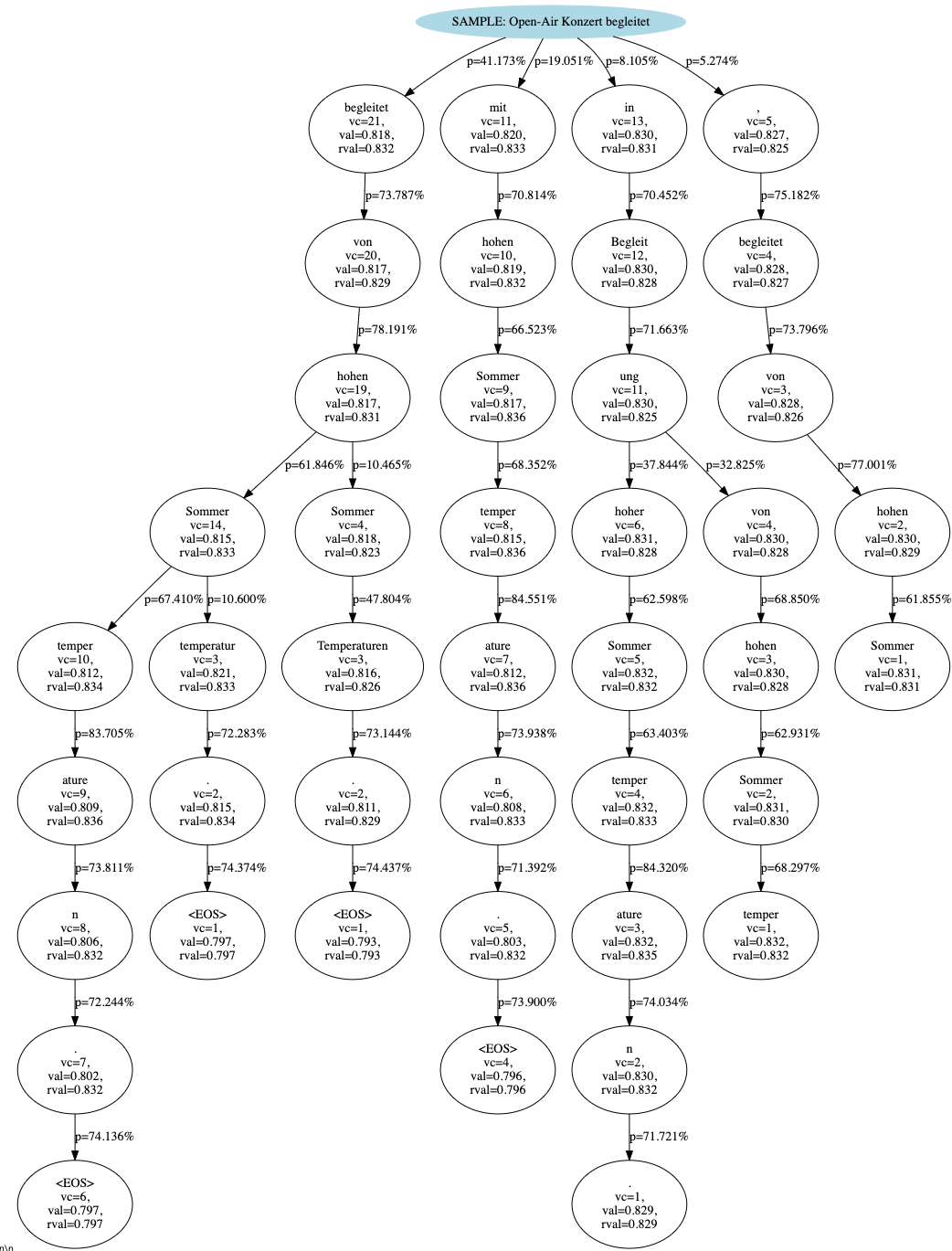}
\caption{MCTS tree when translating ``Open-air concert accompanied by high summer temperatures" from English to German, sixth step.}
\end{figure}

\begin{figure}[H]
\centering
\includegraphics[scale=0.4]{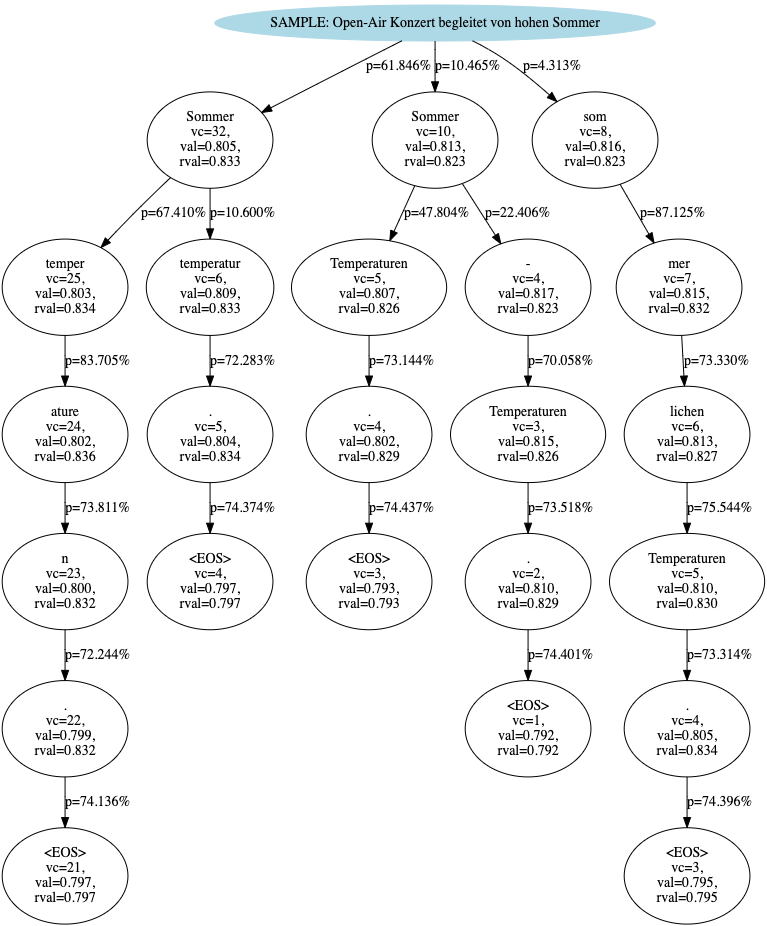}
\caption{MCTS tree when translating ``Open-air concert accompanied by high summer temperatures" from English to German, ninth step.}
\end{figure}

\newpage
\onecolumn
\section{Batched Numpy-friendly MCTS}\label{apx:code}
Accelerator hardware such as GPUs or TPUs allow us to execute neural networks faster; but to fully leverage their computing power, we have to run on batches of several inputs.
This is not very easily mixed with an algorithm such as MCTS, as it requires a queuing mechanism between the search itself and the neural network computations, potentially leading to inefficiencies.
To circumvent this issue, we introduce a Numpy-compatible version of MCTS, which can then be run completely on the accelerator device.

The basic idea is that we use storage tensors which are indexed by the number of the current node or simulation in the MCTS tree.
The root node has index 0 for all elements in a batch, and we then build all subsequent elements recursively.

We start by creating a \texttt{NumpyMCTS} object, whose fields store all the necessary tree information to compute a single batched instance of search (i.e. MCTS for one token, not MCTS applied to the full sequence).
In details, for each node, for each item in the batch we store:

\begin{itemize}
    \item \texttt{visit\_counts}: the amount of times said nodes have been visited during the search,
    \item \texttt{raw\_values}: the initial value of the node as returned by our value network,
    \item \texttt{values}: the aggregated value of the node at this point in the search,
    \item \texttt{parents}: which node is its parent in the tree,
    \item \texttt{action\_from\_parents}: which action was taken to transition from the parent to the node itself,
    \item \texttt{depth}: the tree depth of each node in the tree,
    \item \texttt{is\_terminal}: whether or not they are a terminal node.
\end{itemize}

All these variables are tensors are of size $(B, S)$ where $B$ is the batch size and $S$ is the amount of simulation plus one.

For ease of tree manipulation, we also store for each node the indices of its children, its prior over its possible children, the values of each child, and the visit count of each child.
The associated tensors should be of shape $(B, S, V)$, where V is the total number of possible actions.
However this makes for large tensors, on which Numpy operations can become costly.
To alleviate this issue we store a sparse version of these tensors instead, only keeping the top $A$ children according to the policy for each node.
The shapes are thus $(B, S, A)$ instead.
We maintain a mapping from 0 to $A - 1$ in the \texttt{topk\_mapping} tensor, of shape $(B, S, A)$ itself too.

Finally, the object also stores for each node its associated transformer state, so that we can use incremental inference during the search.
These states can be kept on the accelerator device itself.

\begin{lstlisting}
class NumpyMCTS():

  def __init__(self, root_fun, rec_fun, batch_size, num_simulations, num_actions, num_sparse_actions, pb_c_init):
    self._batch_size = batch_size
    self._num_simulations = num_simulations
    self._num_actions = num_actions
    self._num_sparse_actions = min(num_sparse_actions, num_actions)
    self._pb_c_init = pb_c_init

    self._root_fun = root_fun  # a function called at the root
    self._rec_fun = rec_fun  # a function called in the tree
    self._adaptive_min_values = np.zeros(batch_size, dtype=np.float32)
    self._adaptive_max_values = np.zeros(batch_size, dtype=np.float32)

    # Allocate all necessary storage.
    # For a given search associated to a batch-index, node i is the i-th node
    # to be expanded. Node 0 corresponds to the root node.
    num_nodes = num_simulations + 1
    batch_node = (batch_size, num_nodes)
    self._num_nodes = num_nodes
    self._visit_counts = np.zeros(batch_node, dtype=np.int32)
    self._values = np.zeros(batch_node, dtype=np.float32)
    self._raw_values = np.zeros(batch_node, dtype=np.float32)
    self._parents = np.zeros(batch_node, dtype=np.int32)
    # action_from_parents[b, i] is the action taken to reach node i.
    # Note that action_from_parents[b, 0] will remain -1, as we do not know,
    # when doing search from the root, what action led to the root.
    self._action_from_parents = np.zeros(batch_node, dtype=np.int32)
    # The 0-indexed depth of the node. The root is the only 0-depth node.
    # The depth of node i, is the depth of its parent + 1.
    self._depth = np.zeros(batch_node, dtype=np.int32)
    self._is_terminal = np.full(batch_node, False, dtype=np.bool)

    # To avoid costly numpy ops, we store a sparse version of the actions.
    # We select the top k actions according to the policy, and keep a mapping
    # of indices from 0 to k-1 to the actual action indices in the
    # self._topk_mapping tensor.
    batch_node_action = (batch_size, num_nodes, self._num_sparse_actions)
    self._topk_mapping = np.zeros(batch_node_action, dtype=np.int32)
    self._children_index = np.zeros(batch_node_action, dtype=np.int32)
    self._children_prior = np.zeros(batch_node_action, dtype=np.float32)
    self._children_values = np.zeros(batch_node_action, dtype=np.float32)
    self._children_visits = np.zeros(batch_node_action, dtype=np.int32)
    self._states = {}
    self._batch_range = np.arange(batch_size)
    self._reset_tree()

  def _reset_tree(self):
    """Resets the tree arrays."""
    self._visit_counts.fill(0)
    self._values.fill(0)
    self._parents.fill(-1)
    self._action_from_parents.fill(-1)
    self._depth.fill(0)

    self._topk_mapping.fill(-1)
    self._children_index.fill(-1)
    self._children_prior.fill(0.0)
    self._children_values.fill(0.0)
    self._children_visits.fill(0)
    self._states = {}  # Indexed by tuples (batch index, node index)
\end{lstlisting}

We now define a method to perform the search itself.
As stated in the main text, MCTS consists in applying the same three steps for each simulation, so we iterate over $S$.
First, we use the \texttt{simulate()} method to select which new nodes to explore.
Second, we expand these new nodes (calling our neural network to compute both the policy and the value at these nodes).
Finally, we back the newly computed values up the tree.

The \texttt{dense\_visit\_counts} method allows us to map back our sparse action representation into the original action space.
\begin{lstlisting}
  def search(self, raw_states):
    self._reset_tree()

    # Evaluate the root.
    prior, values, states = self._root_fun(raw_states)

    self._adaptive_min_values = values
    self._adaptive_max_values = values + 1e-6

    root_index = 0
    self._create_node(root_index, prior, values, states, np.full(self._batch_size, False, dtype=np.bool))

    # Do simulations, expansions, and backwards.
    leaf_indices = np.zeros((self._batch_size), np.int32)
    for sim in range(self._num_simulations):
      node_indices, actions = self.simulate()
      next_node_index = sim + 1  # root is 0, therefore we offset by 1.
      self.expand(node_indices, actions, next_node_index)
      leaf_indices.fill(next_node_index)
      self.backward(leaf_indices)

    return self.dense_visit_counts()

  def dense_visit_counts(self):
    root_index = 0
    root_visit_counts = self._children_visits[:, root_index, :]
    dense_visit_counts = np.zeros((self._batch_size, self._num_actions))
    dense_visit_counts[self._batch_range[:, None], self._topk_mapping[:, root_index, :]] = root_visit_counts
    return dense_visit_counts
\end{lstlisting}

The simulate method consists in applying the UCT formula recursively until we have reached a new node to open for each element of the batch.
The UCT formula itself can be computed in a fully batched fashion, as demonstrate by method \texttt{uct\_select\_action}.
\begin{lstlisting}
  def simulate(self):
    """Goes down until all elements have reached unexplored actions."""
    node_indices = np.zeros((self._batch_size), np.int32)
    depth = 0
    while True:
      depth += 1
      actions = self.uct_select_action(node_indices)
      next_node_indices = self._children_index[self._batch_range, node_indices, actions]
      is_unexplored = next_node_indices == -1
      if is_unexplored.all():
        return node_indices, actions
      else:
        node_indices = np.where(is_unexplored, node_indices, next_node_indices)
        
  def uct_select_action(self, node_indices):
    """Returns the action selected for a batch of node indices of shape (B)."""
    node_children_prior = self._children_prior[self._batch_range, node_indices, :]  # (B, A)
    node_children_values = self._children_values[self._batch_range, node_indices, :]  # (B, A)
    node_children_visits = self._children_visits[self._batch_range, node_indices, :]  # (B, A)
    node_visits = self._visit_counts[self._batch_range, node_indices]  # (B)

    node_policy_score = np.sqrt(node_visits[:, None]) * self._pb_c_init * node_children_prior / (node_children_visits + 1)  # (B, A)

    # Remap values between 0 and 1.
    node_value_score = node_children_values
    node_value_score = (node_value_score != 0) * node_value_score + (node_value_score == 0) * self._adaptive_min_values[:, None]
    node_value_score = (node_value_score - self._adaptive_min_values[:, None]) / (self._adaptive_max_values[:, None] - self._adaptive_min_values[:, None])

    node_uct_score = node_value_score + node_policy_score  # (B, A)
    actions = np.argmax(node_uct_score, axis=1)
    return actions
\end{lstlisting}
  
Once we have selected nodes to expand, we can proceed.
The \texttt{expand} method is where we call our neural networks to compute policies and values.
We then create the nodes in the object fields through the \texttt{create\_node} method.
Finally, we update the tree topology to connect the new nodes to the tree.
\begin{lstlisting}        
  def expand(self, node_indices, actions, next_node_index):
    """Creates and evaluate child nodes from given nodes and unexplored actions."""

    # Retrieve states for nodes to be evaluated.
    states = [self._states[(b, n)] for b, n in enumerate(node_indices)]
    previous_node_is_terminal = self._is_terminal[self._batch_range, node_indices[self._batch_range]]  # (B)

    # Convert sparse actions to dense actions for network computation
    dense_actions = self._topk_mapping[self._batch_range, node_indices, actions]

    # Evaluate nodes.
    (prior, values, next_states, expanded_node_is_terminal) = self._rec_fun(states, dense_actions)

    # Create the new nodes.
    self.create_node(next_node_index, prior, values, next_states, expanded_node_is_terminal)

    # Update the min and max values arrays
    self._adaptive_min_values = np.minimum(self._adaptive_min_values, values)
    self._adaptive_max_values = np.maximum(self._adaptive_max_values, values)

    # Update tree topology.
    self._children_index[self._batch_range, node_indices, actions] = next_node_index
    self._parents[:, next_node_index] = node_indices
    self._action_from_parents[:, next_node_index] = actions
    self._depth[:, next_node_index] = self._depth[self._batch_range, node_indices] + 1
    
  def create_node(self, node_index, prior, values, next_states, expanded_node_is_terminal):
    # Truncate the prior to only keep the top k logits
    prior_topk_indices = np.argpartition(prior, -self._num_sparse_actions, axis=-1)[:, -self._num_sparse_actions:]
    prior = prior[self._batch_range[:, None], prior_topk_indices]  # (B, A)

    # Store the indices of the top k logits
    self._topk_mapping[self._batch_range, node_index, :] = prior_topk_indices

    # Update prior, values and visit counts.
    self._children_prior[:, node_index, :] = prior
    self._values[:, node_index] = values
    self._raw_values[:, node_index] = values
    self._visit_counts[:, node_index] = 1
    self._is_terminal[:, node_index] = expanded_node_is_terminal

    # Update states.
    for b, next_state in enumerate(next_states):
      self._states[(b, node_index)] = next_state

\end{lstlisting}

Finally, we back the newly computed values up using the \texttt{backward} method, which can again be done in a fully batched fashion.
\begin{lstlisting}
  def backward(self, leaf_indices):
    """Goes up and updates the tree until all nodes reached the root."""
    node_indices = leaf_indices  # (B)
    leaf_values = self._values[self._batch_range, leaf_indices]
    while True:
      is_root = node_indices == 0
      if is_root.all():
        return
      parents = np.where(is_root, 0, self._parents[self._batch_range, node_indices])
      root_mask = 1.0 * is_root
      not_root_mask_int = (1 - is_root)
      not_root_mask = 1.0 - root_mask
      # Update the parent nodes iff their child is not the root.
      # We therefore mask the updates using not_root_mask and root_mask.
      self._values[self._batch_range, parents] = not_root_mask * (self._values[self._batch_range, parents] * self._visit_counts[self._batch_range, parents] + leaf_values) / (self._visit_counts[self._batch_range, parents] + 1.0) + root_mask * self._values[self._batch_range, parents]
      self._visit_counts[self._batch_range, parents] += not_root_mask_int
      actions = np.where(is_root, 0, self._action_from_parents[self._batch_range, node_indices])
      self._children_values[self._batch_range, parents, actions] = not_root_mask * self._values[self._batch_range,node_indices] + root_mask * self._children_values[self._batch_range, parents, actions]
      self._children_visits[self._batch_range, parents, actions] += not_root_mask_int

      # Go up
      node_indices = parents
\end{lstlisting}

\end{document}